%% file: main.tex
\documentclass[10pt,twocolumn,letterpaper]{article}

\usepackage[pagenumbers]{cvpr} 

\input{preamble}
\definecolor{cvprblue}{rgb}{0.21,0.49,0.74}
\usepackage[pagebackref,breaklinks,colorlinks,allcolors=cvprblue]{hyperref}

\def\paperID{5167} 
\def\confName{CVPR}
\def\confYear{2026}

\title{UniGS: Unified Geometry-Aware Gaussian Splatting for Multimodal Rendering}

\author{
Yusen Xie$^{1}$ \quad  Zhenmin Huang$^{2}$ \quad  Jianhao Jiao$^{3}$ \quad Dimitrios Kanoulas$^{3}$ \quad Jun Ma$^{1,2}$\\
$^{1}$HKUST (GZ)  \quad $^{2}$HKUST \quad $^{3}$UCL \\
\tt\small \{yxie827@connect.hkust-gz.edu.cn; zhuangdf@connect.ust.hk; ucacjji@ucl.ac.uk;\\ \tt\small d.kanoulas@ucl.ac.uk; jun.ma@ust.hk\}
\\
\tt\small Project Page: \url{
https://github.com/xieyuser/UniGS
}
}

\usepackage{pifont}
\usepackage{multirow}
\usepackage[table,xcdraw]{xcolor}
\usepackage[linesnumbered,ruled,vlined]{algorithm2e}
\usepackage{algpseudocode}
\usepackage{xr}
\usepackage{soul}
\usepackage[normalem]{ulem}
\useunder{\uline}{\ul}{}

\usepackage{etoolbox}

\makeatletter
\let\old@algocf@start\@algocf@start
\renewcommand{\@algocf@start}{%
  \vspace{-1em}%
  \old@algocf@start%
}

\let\old@algocf@finish\@algocf@finish
\renewcommand{\@algocf@finish}{%
  \old@algocf@finish%
  \vspace{-1em}%
}
\makeatother

\definecolor{tabfirst}{rgb}{1, 0.7, 0.7} 
\definecolor{tabsecond}{rgb}{1, 0.85, 0.7} 
\definecolor{tabthird}{rgb}{1, 1, 0.7} 

\newcommand{\mytabref}[1]{Tab.~\ref{#1}}
\newcommand{\myeqref}[1]{(\ref{#1})}
\newcommand{\myfigref}[1]{Fig.~\ref{#1}}
\newcommand{\myalgref}[1]{Alg.~\ref{#1}}
\newcommand{\mysecref}[1]{Sec.~\ref{#1}}

\expandafter\def\expandafter\normalsize\expandafter{%
    \normalsize%
    \setlength\abovedisplayskip{4pt}%
    \setlength\belowdisplayskip{4pt}%
}

\newcommand{\mysection}[1]{%
    \vspace{-0.1em}%
    \section{#1}%
    \vspace{-0.1em}%
}

\newcommand{\mysubsection}[1]{%
    \vspace{-0.1em}%
    \subsection{#1}%
    \vspace{-0.1em}%
}

\algnewcommand\algorithmicswitch{\textbf{switch}}
\algnewcommand\algorithmiccase{\textbf{case}}
\algnewcommand\algorithmicassert{\texttt{assert}}
\algnewcommand\Assert[1]{\State \algorithmicassert(#1)}%
\algdef{SE}[SWITCH]{Switch}{EndSwitch}[1]{\algorithmicswitch\ #1\ \algorithmicdo}{\algorithmicend\ \algorithmicswitch}%
\algdef{SE}[CASE]{Case}{EndCase}[1]{\algorithmiccase\ #1}{\algorithmicend\ \algorithmiccase}%
\algtext*{EndSwitch}%
\algtext*{EndCase}%

\begin{document}
\twocolumn[{
\maketitle
 \begin{center}
 \centering
 \vspace{-0.3in}
 \includegraphics[width=\linewidth]{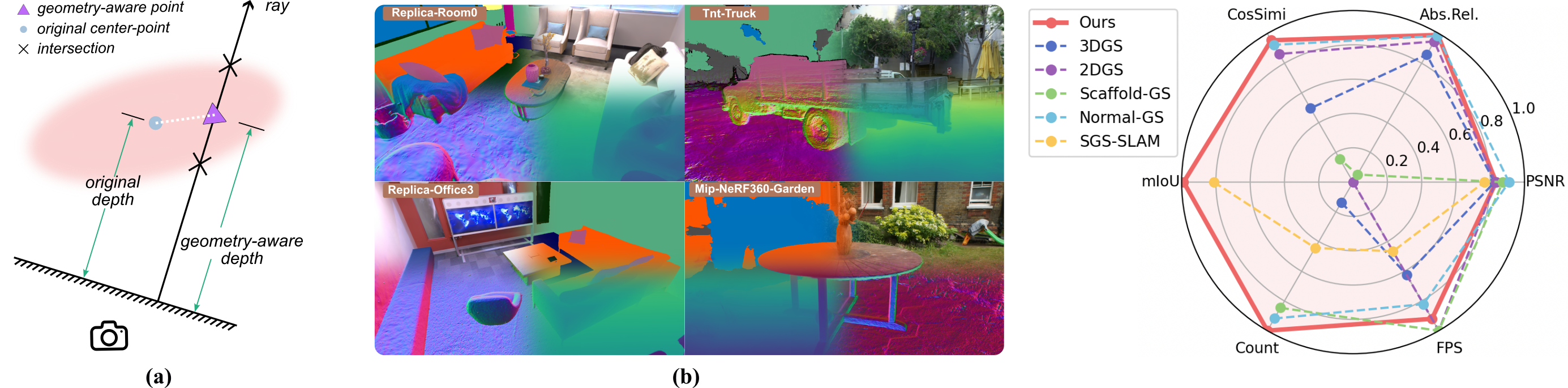}
     \captionof{figure}{We propose a unified 3D Gaussian Splatting framework that jointly predicts RGB, depth, normal, and semantic map through a single forward rasterization process. (a) Our geometry-aware rendering method explicitly incorporates rotation and scaling attribute into both the rasterization and gradient propagation processes, where we derive an analytical solution for gradient propagation to significantly accelerate optimization. (b) We present multimodal rendering results across diverse scenes, showcasing RGB images, depth maps, surface normals, and semantic segmentation map in a four-panel visualization layout. (c) A comprehensive comparison with widely-used baselines in \textit{Replica-Room0} dataset, underscoring the full modality coverage and strong performance of our framework. All scores are normalized to the [0, 1] range by defining the worst and best reference values for each metric individually, ensuring 1 represents optimal performance. Incomplete dashed lines indicate that the corresponding baseline does not support specific modalities. Evaluation covers PSNR for RGB quality, absolute relative error (\textit{Abs.Rel.}) for depth estimation, cosine similarity (CosSimi) for normal estimation, mean intersection-over-union (mIoU) for semantic segmentation, as well as system performance metrics including the number of Gaussian primitives (Count) and rendering speed in frames per second (FPS).}
     \label{fig:firstdemo}
 \end{center}
}]

\input{sec/0_abstract}    
\input{sec/1_intro} 
\input{sec/2_related}
\input{sec/3_method}

\input{sec/4_exps}
\input{sec/6_conclusion}

\input{sec/X_suppl}

{
    \small
    \bibliographystyle{ieeenat_fullname}
    \bibliography{main}
}

\end{document}

%% file: sec/0_abstract.tex
\begin{abstract}
In this paper, we propose UniGS, a unified map representation and differentiable framework for high-fidelity multimodal 3D reconstruction based on 3D Gaussian Splatting. Our framework integrates a CUDA-accelerated rasterization pipeline capable of rendering photo-realistic RGB images, geometrically accurate depth maps, consistent surface normals, and semantic logits simultaneously. We redesign the rasterization to render depth via differentiable ray-ellipsoid intersection rather than using Gaussian centers, enabling effective optimization of rotation and scale attribute through analytic depth gradients. Furthermore, we derive the analytic gradient formulation for surface normal rendering, ensuring geometric consistency among reconstructed 3D scenes. To improve computational and storage efficiency, we introduce a learnable attribute that enables differentiable pruning of Gaussians with minimal contribution during training. Quantitative and qualitative experiments demonstrate state-of-the-art reconstruction accuracy across all modalities, validating the efficacy of our geometry-aware paradigm. Source code and multimodal viewer will be available on GitHub.
\end{abstract}

%% file: sec/1_intro.tex
\mysection{Introduction}
\label{sec:intro}

High-quality novel view synthesis (NVS) and fast multimodal 3D reconstruction remains a challenging problem in computer graphics and attracts widespread attention in fields such as autonomous driving~\cite{yan2024street, huang2025gaussianformer,huang2024gaussianformer,zheng2024gaussianad} and robotics~\cite{zhu20243d,lu2024manigaussian}. 
In recent years, neural radiance fields (NeRF)~\cite{verbin2022ref,barron2023zip,barron2022mip,Mildenhall20eccv_nerf} and 3D Gaussian Splatting (3DGS)~\cite{kerbl2024hierarchical,3dgs} have made big breakthroughs in this field. NeRF uses coordinate-based networks to store 3D scene in multi-layer perceptrons (MLPs) via ray sampling, bringing in photo-realistic reconstruction. 3DGS, another approach, uses explicit 3D Gaussian primitives and modern GPU-accelerated splatting, and this design has been shown to enable high-quality real-time NVS by many 3DGS-based frameworks~\cite{scaffoldgs,lee2024compact,fan2024lightgaussian,niedermayr2024compressed,huang20242d,xie2024gs, ren2024octree, yan2024street,liu2024citygaussian,liu2024citygaussianv2}. However, the two types of existing methods still have many issues, which mainly focus on the following aspects.
NeRF requires long training times, lacks easy scene editability, and its implicit 3D information storage method renders geometric reconstruction less intuitive~\cite{semanticnerf,Mildenhall20eccv_nerf}. Additionally, it performs poorly when applied to large-scale scenes~\cite{shi2025normal,blocknerf}. 3DGS and its related frameworks have more challenges: 1)
Poor geometric structure and consistency: conventional image-centric 3DGS often ignores objects’ geometric structure and semantic info. It only uses Gaussian centers to represent depth, leaving out Gaussian rotation and scale attributes. This stops gradient propagation for geometric constraints (e.g. depth and normal), leading to inconsistent surface normals~\cite{scaffoldgs,3dgs,li2024geogaussian,normalgs} and misalignment between Gaussians primitives and reconstructed surfaces.
2) Inefficient densification and pruning: some 3DGS methods create too many redundant Gaussians primitives to fit RGB images well~\cite{kerbl2024hierarchical,li2024geogaussian,3dgs,yan2024street}. Though this improves metrics, it slows down rendering and uses more GPU memory, losing 3DGS’s real-time edge.
3) Poor unified multimodal framework: some 3DGS-related methods use separate frameworks for multimodal rendering~\cite{sgsslam, guo2024semantic}, which causes repeated calculations and inconsistent 3D scene representations, failing to unify multimodal modeling.
Existing studies have tried single or multiple scene modalities~\cite{scaffoldgs, normalgs, li2024geogaussian, deng2025omnimap}, but there is still no unified framework that combines all these modalities while keeping storage efficient and rendering fast. This gap serves as the key motivation for our approach.


Against this backdrop, we propose \textbf{UniGS}, a \textit{unified} geometry-aware 3D Gaussian Splatting map representation and framework that establishes a new paradigm for consistent 3D scene reconstruction. During rendering, each pixel is shaded via backward ray tracing to determine its intersection with individual Gaussian ellipsoids (\myfigref{fig:firstdemo}(a)). This formulation explicitly integrates Gaussian rotation and scale attributes into the rendering pipeline. By leveraging analytical gradients derived from depth and normal supervision, it enables efficient gradient-based optimization of these geometric attributes.
Furthermore, we introduce a learnable attribute to continuously prune Gaussians. Experimental results demonstrate that this attribute significantly improves computational and storage efficiency without sacrificing rendering or geometric accuracy.
Crucially, our framework achieves high-fidelity rendering of \textit{RGB, depth, normals, and semantic logits} through a single differentiable pipeline simultaneously (\myfigref{fig:firstdemo}(b)). The loss from each modality can be back-propagated via analytical gradients to the shared 3D Gaussian representation, thereby maintaining consistency across all modalities in the reconstructed environment (\myfigref{fig:firstdemo}(c)). Particularly, our method achieves a 66.4\% improvement in depth estimation accuracy while reducing the Gaussian primitive count by 17.2\%.
In this paper, our contributions are summarized as follows:
\begin{itemize}
\item We develop a novel differentiable depth rasterization method based on ray-ellipsoid intersection, which enables optimization of Gaussian rotation and scale attributes via analytic gradients propagation. This approach ensures that Gaussian primitives conform closely to underlying surface geometries, significantly improving geometric consistency in reconstructed scenes.
\item We propose a trainable gradient factor in map representation for dynamically pruning insignificant Gaussians in a differentiable manner, which enhances rendering speed and storage efficiency.
\item We propose a unified map representation and differentiable rasterization framework that integrates multimodal data. This integration not only enables mutual enhancement across modalities but also improves rendering efficiency and strengthens the consistency of the reconstructed environment.
\item Through comprehensive experimental evaluation, we demonstrate that our approach achieves state-of-the-art reconstruction quality across all modalities while consistently maintaining real-time performance. 
\end{itemize}

%% file: sec/2_related.tex
\begin{figure*}[!h]
\centering
\includegraphics[width=\linewidth]{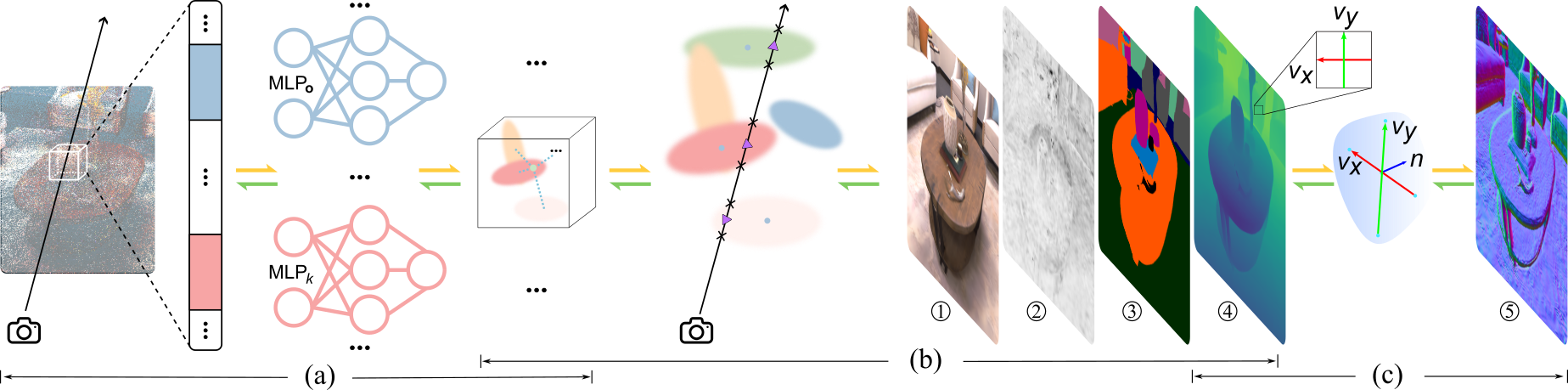}
\caption{\textbf{Overview of our framework}. The yellow arrows ($\textcolor[HTML]{ffcc3d}{\rightarrow}$) indicate the forward rasterization pipeline. (a) First, we predict the semantic logits $\textbf{o}$, contribution $k$, and other 3DGS attributes (i.e., position $\boldsymbol\mu$, rotation $\textbf{q}$, scale $\textbf{s}$, opacity $\alpha$, SHs $\textbf{h}$) in the anchor set through MLPs. For each anchor, we predict $M$ raw 3DGS data available for rendering. (b) Then, during the rasterization stage, for each pixel, we use ray tracing to calculate its intersection ($\times$) with the Gaussian ellipsoid, which is closely related to the rotation and scale attributes. We replace the depth of the Gaussian center point (\textcolor[HTML]{7dc2da}{$\bullet$}) with the depth of the midpoint (\textcolor[HTML]{be6af9}{$\blacktriangle$}) between the two intersection points. After splatting rasterization, we obtain RGB (\ding{172}), gradient factor map (\ding{173}), semantic (\ding{174}), and depth (\ding{175}). (c) To obtain the surface normal (\ding{176}), we back-project the depth into three-dimensional space and then use a differential method to obtain the normals in the world coordinate system, ultimately producing the normal rendering map. The final rendered results are compared with their respective ground truths to calculate the loss. The green arrows ($\textcolor[HTML]{80cb60}{\leftarrow}$) indicate the direction of gradient propagation through a CUDA-accelerated analytical solution.}
\label{fig:overview}
\vspace{-1em}
\end{figure*}

\vspace{-1em}
\mysection{Related Works}
\noindent\textbf{Photo-Realistic Rendering and Densification.}
In the domain of photo-realistic reconstruction and rendering, NeRF and 3DGS represent implicit and explicit techniques, respectively. NeRF \cite{Mildenhall20eccv_nerf,barron2022mip} and its variants pioneer the use of implicit neural representations for NVS, achieving high-fidelity rendering by optimizing a continuous volumetric scene network using multi-view images. However, this approach typically requires extremely long training times. 
Subsequent works incorporate geometric prior structures~\cite{yu2021plenoxels, barron2023zip,tang2024hisplat} or efficient encoding~\cite{muller2022instant} to accelerate the process. Despite these improvements, NeRF still faces inherent limitations, such as the difficulty in editing implicitly represented scenes. 
In contrast, 3DGS \cite{3dgs,yang2024spectrally} employs explicit Gaussian primitives as scene representations. 
Through compression~\cite{scaffoldgs,lee2024compact,fan2024lightgaussian,niedermayr2024compressed,huang20242d} and efficient geometric structures~\cite{kerbl2024hierarchical,chen2024hac}, several studies extend 3DGS to large-scale outdoor~\cite{xie2024gs, ren2024octree, yan2024street, li2025scenesplat} and urban~\cite{liu2024citygaussian,liu2024citygaussianv2} environments while maintaining high-quality reconstruction. But the aforementioned research primarily focuses on RGB-based rendering and seldom incorporates explicit geometric information of the environment. Moreover, impressive efforts have been made to enhance optimization and storage efficiency. SteepestGS~\cite{wang2025steepest} uncovers fundamental principles of density control. DashGaussian~\cite{chen2025dashgaussian} significantly accelerates 3DGS optimization. GeoTexDensifier~\cite{jiang2024geotexdensifier} utilizes geometric information for Gaussian densification, and Pixel-GS~\cite{zhang2024pixel} introduces pixel-level regularization for density control. These methods are regarded as post-processing techniques and lack integration within a unified framework.

\noindent\textbf{Geometry-Aware Rendering.} 
Geometry-aware approaches enhance reconstruction accuracy by incorporating surfel~\cite{Huang20242DGS,normalgs,jiang2024gaussianshader,guedon2024sugar,dai2024high}, normals~\cite{normalgs,shi2025normal}, and relighting~\cite{gao2024relightable,wu2024deferredgs,ye20243d} capabilities, enabling more physically realistic renderings and interactions. Some studies~\cite{wang2021neus,oechsle2021unisurf,yariv2021volume,shi2025normal} attempt to achieve accurate geometry within the NeRF framework, and while significant progress has been made, these methods still suffer from inherent limitations that are difficult to overcome with implicit representations. 3DGS~\cite{3dgs,scaffoldgs} does not inherently excel in geometric reconstruction either, as its Gaussian primitives are independent of each other, making precise geometry recovery an open challenge. Several works unitize 3DGS or 2DGS to recover depth~\cite{3dgs,scaffoldgs,ververas2024sags,normalgs}, surfel~\cite{Huang20242DGS,normalgs,jiang2024gaussianshader} and meshes~\cite{Huang20242DGS,guedon2024sugar,dai2024high}, including normal estimation~\cite{normalgs} and relighting~\cite{gao2024relightable,wu2024deferredgs,ye20243d} techniques, which can effectively reconstruct smooth object surfaces. Nevertheless, these approaches remain largely limited to small-scale objects. When extended to large environments such as outdoor or urban scenes~\cite{loccoz20243dgrt,wu20253dgut}, the required computational time remains substantial. 
Some methods attempt to incorporate SDF-based surfel constraints~\cite{yu2024gsdf,yariv2021volume,wang2023neus2,li2023neuralangelo,yariv2023bakedsdf}, inverse rendering~\cite{shi2025gir,ye2024geosplating}, or restrictions~\cite{li2024geogaussian,Keetha2023SplaTAMST} on Gaussian properties. However, these approaches lack rigorous analytical derivation for geometric modeling, employ relatively limited quantitative metrics, and demonstrate constrained generalizability across diverse scenarios. Consequently, the improvements achieve thus far remain marginal in terms of geometric fidelity.

\noindent\textbf{Semantic Rendering Frameworks.}
Several works attempt to integrate semantic information into both NeRF~\cite{semanticnerf,chou2024gsnerf} and 3DGS~\cite{guo2024semantic,qin2024langsplat}, including techniques like Open-Vocabulary Segmentation~\cite{zhang2023simple, lai2025exploring,deng2025omnimap} and Language-Driven Semantic Segmentation~\cite{radford21a,liang2023open,luo2023segclip,li2022language}, with promising results. Embedding semantic information facilitates downstream tasks such as scene understanding~\cite{jiang2025votesplat,zhou2024hugs}, navigation~\cite{lei2025gaussnav,jin2024gs}, and scene editing~\cite{ye2024gaussian,3dgsdrag2025}. For instance, SGS-SLAM~\cite{sgsslam} directly uses semantic ground truth as supervision signal and rendering output in 3DGS, though it fails to formulate a unified 3DGS map representation. Feature-GS~\cite{zhou2024feature} extracts semantic features from the environment using 3DGS, and SEGS-SLAM \cite{wen2025segsslam} enhances structural awareness with appearance embeddings. 
Nevertheless, such approaches exhibit limitations in reconstructing accurate environmental geometry and fail to achieve consistent reconstruction of scene geometry and semantics.


%% file: sec/3_method.tex
\mysection{Methodology}
\label{sec:methodology}
The methodology section is organized as follows. Firstly, we define our map representation and objective of this work (\mysecref{sec:representation}). Then, the subsequent sections detail the key components of our framework: geometry-aware depth rasterization and normal estimation (\mysecref{sec:depth_normal}); the novel gradient factor and differentiable pruning strategy (\mysecref{sec:pruning}); and finally, the overall multimodal rendering pipeline, loss functions, and implementation details (\mysecref{sec:framework}). The brief overview of our framework is illustrated in~\myfigref{fig:overview}.

\mysubsection{Unified Map Representation and Objectives}
\label{sec:representation}
\noindent\textbf{Unified Map Representation}. In our framework, each Gaussian $\mathcal{G}$ is defined by position $\boldsymbol{\mu} \in \mathbb{R}^{3}$, quaternion $\mathbf{q} \in \mathbb{R}^{4}$, scale $\mathbf{s} \in \mathbb{R}^{3}$, opacity $\alpha \in \mathbb{R}$, and Spherical Harmonics (SHs) $\mathbf{h} \in \mathbb{R}^{C_{\text{h}}}$ ($C_{\text{h}}$ is the pre-difined SHs number per color channel), semantic logits (encode the probability distribution across $C_{\text{o}}$ semantic categories) $\mathbf{o} \in \mathbb{R}^{C_{\text{o}}}$
and gradient factor $k \in \mathbb{R}$:
\begin{equation}
\label{eq:gs_representation}
\mathcal{G} =  \{\boldsymbol{\mu},\,\mathbf{q},\,\mathbf{s},\,\alpha,\,\mathbf{h},\,\mathbf{o},\,k\}.
\end{equation}
\noindent\textbf{Objective}. The objective of this framework is to render RGB color image, geometric depth, surface normals, semantic logits, and a gradient factor map (used in differentiable pruning) \textit{simultaneously} within a \textit{unified} rasterization pipeline. These rendered outputs are then compared against their corresponding 2D ground truth to compute respective loss, which guide the optimization of the 3D Gaussian scene through a differentiable backward process. To accelerate convergence and enhance computational efficiency, we derive the relevant \textit{analytical solutions} and implement them using \textit{CUDA-accelerated} gradient propagation. Finally, we can obtain a geometry consistent photo-realistic 3DGS map representation.

\mysubsection{Geometry-Aware Rasterization}
\label{sec:depth_normal}
\noindent\textbf{Ray-Gaussian Primitive Intersection Formulation.}
We define an ellipsoid \(\mathcal{E}\) associated with Gaussian \(\mathcal{G}\) by its center \(\boldsymbol{\mu}\), rotation matrix \(\mathbf{R}\) (converted from quaternion \(\mathbf{q}\)) and diagonal scaling matrix \(\mathbf{S} = \text{diag}(\mathbf{s}_x, \mathbf{s}_y, \mathbf{s}_z)\), and this ellipsoid consists of all 3D points \(\mathbf{x}\) on its surface that satisfy the following equation:
\begin{equation}\left(\mathbf{S}^{-1} \mathbf{R}^{-1} (\mathbf{x} - \boldsymbol{\mu}) \right)^{\top} \left(\mathbf{S}^{-1} \mathbf{R}^{-1} (\mathbf{x} - \boldsymbol{\mu}) \right) = 1.\end{equation}
For each pixel $p = (u, v)$, the corresponding camera ray $\mathbf{r}(t) = \mathbf{t}_{\text{cam}}^{\text{world}} + t\mathbf{d}$ is calculated through a unprojection process:
\begin{equation}
\label{eq:computeRay}
\begin{aligned}
\mathbf{r}(t) = \mathbf{t}_{\text{cam}}^{\text{world}} + \operatorname{normalize}\left( \mathbf{R}_{\text{cam}}^{\text{world}} \begin{bmatrix} \frac{u - c_x}{f_x} \\ \frac{v - c_y}{f_y} \\ 1 \end{bmatrix} \right),
\end{aligned}
\end{equation}
where $\mathbf{t}_{\text{cam}}^{\text{world}}$ and $\mathbf{R}_{\text{cam}}^{\text{world}}$ are the translation vector and rotation matrix that transform a point in the camera frame to the world frame, respectively. The focal lengths $f_x, f_y$ and principal point $c_x, c_y$ are camera parameters.

To find the intersection between the ray $\mathbf{r}(t)$ and the ellipsoid $\mathcal{E}$, we transform the ray $\mathbf{r}(t)$ into the $\textit{local}\,(l)$ $\textit{scaled}\,(s)$ coordinate system of $\mathcal{E}$:
\begin{equation}
\begin{aligned}
\mathbf{v}_{s} &= \mathbf{S}^{-1} \mathbf{v}_{l},\quad\mathbf{v}_{l} = \mathbf{R}^{-1} (\mathbf{t}_{\text{cam}}^{\text{world}} - \boldsymbol{\mu}), \\
\mathbf{d}_{s} &= \mathbf{S}^{-1} \mathbf{d}_{l},\quad\mathbf{d}_{l} = \mathbf{R}^{-1} \mathbf{d}.
\end{aligned}
\end{equation}
The intersection problem is then reduced to solving for $t$ in the unit sphere equation:
\begin{equation}
\label{equ:rayEllipsoidIntersection}
\| \mathbf{v}_{s} + t \mathbf{d}_{s} \|^2 = 1.
\end{equation}
The solution to this equation can be treated as that of a quadratic equation: for a light ray intersecting an ellipsoid, there are two valid solutions ($t_1, t_2$) for two intersection points, $t_1 = t_2$ for one point, and no valid solutions for no intersection. Details are in the supplementary materials.

\noindent\textbf{Depth Rasterization.} We take the \textit{midpoint} $t_{\text{mid}} = (t_1 + t_2)/2$ of the two valid solutions ($t_1$, $t_2$) as the intersection results. The replaced depth $d$ of corresponding 3D primitive in camera coordinate can be computed by
\begin{equation}\label{eq:midpoint}
\begin{aligned}
\mathbf{P}_{\text{cam}} = \mathbf{R}_{\text{world}}^{\text{cam}} (\mathbf{t}_{\text{cam}}^{\text{world}} + t_{\text{mid}} \mathbf{d}),\quad d = \mathbf{P}_{\text{cam},\,z}.
\end{aligned}
\end{equation}
The final depth value $\tilde{D}(u, v)$ for pixel $p = (u, v)$ is obtained by alpha-rendering the depths $d$ of all overlapping Gaussians $\mathcal{N}$:
\begin{equation}
\label{equ:depth_render}
\tilde{\mathbf{D}}(u, v) =  \sum_{i\in \mathcal{N}}^{} \alpha_i d_i \prod_{j=1}^{i-1}(1 - \alpha_j).
\end{equation}
The above method allows to directly incorporate the geometric information (rotation $\mathbf{q}$ and scale $\mathbf{s}$ of the Gaussian $\mathcal{G}$) into the rasterization pipeline. This enables us to compute the gradients of the loss function with respect to the rotation $\mathbf{q}$ and scale $\mathbf{s}$, thereby effectively optimizing the Gaussian $\mathcal{G}$. Related gradient backpropagation derivations are provided in the supplementary materials.

\begin{algorithm}[htbp]
\caption{Surface Normal Estimation by Depth}
\label{alg:normal_estimation}
\textbf{Input:} Depth $\tilde{\mathbf{D}}\,(u, v)$,\,\texttt{STEP\_1},\,\texttt{STEP\_2},\,$\lambda$\\
\textbf{Output:} Normal $\tilde{\mathbf{N}}(u, v)$\\
\For{each pixel $p=(u, v)$ in parallel}{
    \textbf{1. Backproject to World Coordinate} \\
    \textbf{2. Compute Multi-Scale Normals} \\
    \Indp
    \For{$g_i \in \{\texttt{STEP\_1}, \texttt{STEP\_2}\}$}{
        Finite differences of neighboring points: $\begin{aligned}
\mathbf{v}_{x,i} &= \mathbf{P}(u + g_i, v) - \mathbf{P}(u - g_i, v),\\ \quad
\mathbf{v}_{y,i} &= \mathbf{P}(u, v + g_i) - \mathbf{P}(u, v - g_i),\
\end{aligned}$ \\
        Estimate normal: $\mathbf{n}_{i} = \mathbf{v}_{x,i} \times \mathbf{v}_{y,i}$
    }
    \Indm
    \textbf{3. Fuse and Normalize} \\
    \Indp
    Ensure consistency: $\mathbf{n}_2 \leftarrow \mathbf{n}_2 \cdot \text{sign}(\mathbf{n}_1 \cdot \mathbf{n}_2)$ \\
    Fuse estimates: $\mathbf{n}_{\text{fused}} = \lambda \mathbf{n}_1 + (1-\lambda)\mathbf{n}_2$ \\
    Normalize: $\tilde{\mathbf{N}}(u, v) = \mathbf{n}_{\text{fused}} / \|\mathbf{n}_{\text{fused}}\|$ \\ 
    \Indm
    \textbf{4. Orient Towards View Direction} \\
}
\end{algorithm}

\noindent\textbf{Normal Rasterization from Depth.} As mentioned previously, it is challenging to obtain sufficiently smooth and accurate surface normals through rasterization-based rendering due to the strong directional information inherent in normals. Therefore, in this paper, we compute normals indirectly from depth rasterization result. The focus lies on deriving the gradient propagation process from the normal loss back to the depth. 

Our approach begins by unprojecting the depth map into world coordinate using the camera parameters. We then estimate the normals in world coordinate through a multi-step finite difference scheme, with step sizes denoted as $\texttt{STEP\_1}, \texttt{STEP\_2}$. The overall procedure is summarized in~\myalgref{alg:normal_estimation}. Further details regarding the backprojection of depth pixels and orientation toward the view direction will be provided in the supplementary material. Additionally, the derivation of backpropagation from the normal loss to depth will also be included in the supplementary material.

\mysubsection{Differentiable Pruning}\label{sec:pruning}
\noindent\textbf{Gradient Factor Rasterization and Backpropagation}. We define a new attribute $k$ in~\myeqref{eq:gs_representation} for each Gaussian $\mathcal{G}$ to measure its gradient level during the rendering process. Alongside the rendering of RGB image, we can render a 2D gradient factor map $\tilde{\mathbf{K}}$ by alpha-blending:
\begin{equation}
\tilde{\mathbf{K}}(u, v) = \sum_{i\in \mathcal{N}}^{}  \alpha_i k_i \prod_{j=1}^{i-1}(1 - \alpha_j).
\label{eq:t_render}
\end{equation}
Our objective is to drive each Gaussian’s $k$ toward 1. Experimental results show this state corresponds to the optimal optimization effect of Gaussians. To enforce this, we minimize the L1-loss between $\tilde{\mathbf{K}}$ and an all-ones matrix $\mathbf{1}$:  
\begin{equation}
\mathcal{L}_K = \| \tilde{\mathbf{K}} - \mathbf{1} \|_1.
\end{equation}  
Detailed analysis on why $k = 1$ is optimal and gradient backpropagation are provided in the supplementary materials.

\noindent\textbf{Pruning}. After a fixed interval of iterations $K_p$ ($K_p=3000$ in our experiments), our framework directly eliminates Gaussians with abnormal gradient factor $\| k - \mathbf{1} \|_1 > \mathcal{T}_k$ , effectively pruning the set and reducing the number of Gaussians to enhance overall efficiency. After pruning operation, the $k$ value of all Gaussians will be reset to 0.9 to maintain the sustainability of updates. Please refer to our code for details.

\mysubsection{Multimodal Rendering Framework}
\label{sec:framework}

\noindent\textbf{Architecture Overview}.
Our system is built upon a modified Scaffold-GS~\cite{scaffoldgs} architecture. For each anchor point, we maintain a foundational 3D Gaussian representation~\myeqref{eq:gs_representation}. During every training iteration, multiple MLPs are used to predict attribute offsets for each anchor.

\noindent\textbf{Attribute Prediction}. 
For semantic logits $\mathbf{o} \in \mathbb{R}^{C_{\mathbf{o}}}$ and gradient factor $k \in \mathbb{R}$ of a Gaussian $\mathcal{G}$, we replicate (denoted as $\textbf{Repeat}$) $M$ times and modulate by the output of an MLP branch to obtain Gaussian features for rasterization:
\begin{equation}
 \begin{bmatrix}\mathbf{o}_{\text{vis}}
 \\
k_{\text{vis}}\\...
\end{bmatrix}  = \textbf{Repeat}(\begin{bmatrix}\mathbf{o}_{\text{ah}}
 \\
k_{\text{ah}}\\...
\end{bmatrix}, M) \otimes \sigma\begin{bmatrix}\textbf{MLP}_{\textbf{o}}(\mathbf{f}_{\text{ah}})
 \\\textbf{MLP}_k(\mathbf{f}_{\text{ah}})\\...
\end{bmatrix},
\end{equation}
where $\mathbf{o}_{\text{ah}} \in \mathbb{R}^{C_{\mathbf{o}}}$ is the base semantic logits of the anchor, $k_{\text{ah}} \in \mathbb{R}$ is the base gradient factor of the anchor. $M$ is the replication factor. $\mathbf{f}_{\text{ah}}$ represents the input feature for the anchor. $\text{MLP}_{\text{o}}: \mathbb{R}^{Fdim} \rightarrow \mathbb{R}^{C_{\mathbf{o}}}$ denotes the MLP branch for semantics. $\text{MLP}_{k}: \mathbb{R}^{Fdim} \rightarrow \mathbb{R}$ denotes the MLP branch for gradient factor. $\sigma$ is the sigmoid activation function. $\otimes$ denotes element-wise multiplication. $\mathit{Fdim}$ is the feature dimension.
The resulting modulated attributes are then combined with other predicted properties (i.e., position $\boldsymbol{\mu}$, rotation $\textbf{q}$, scale $\textbf{s}$, opacity $\alpha$, SHs $\textbf{h}$, denoted as $...$) to form the final 3D Gaussians ready for projection and rendering.

\begin{algorithm}[!htb]
\caption{Multimodal Forward Rasterization Pipeline}
\label{alg:forward_render}
\KwIn{$\mathcal{G} =  \{\boldsymbol{\mu},\,\mathbf{q},\,\mathbf{s},\,\alpha,\,\mathbf{h},\,\mathbf{o},\,\mathbf{k}\}$, Image dimensions $W, H$}
\KwOut{$\tilde{\mathbf{C}} \in \mathbb{R}^{W \times H \times 3}$, $\tilde{\mathbf{O}} \in \mathbb{R}^{W \times H \times C}$, $\tilde{\mathbf{D}} \in \mathbb{R}^{W \times H}$, $\tilde{\mathbf{K}} \in \mathbb{R}^{W \times H}$}

\ForPar{each pixel = $(u, v)$}{
    \tcp{Compute Ray of pixel}
    $\text{computeRay}~\myeqref{eq:computeRay}$\;
    Initialize accumulators: $\mathbf{C} \gets \mathbf{0}$, $\mathbf{O} \gets \mathbf{0}$, $D \gets 0$, $K \gets 0$, $T \gets 0$\;
    
    \For{each Gaussian $i$ intersecting pixel $(u, v)$}{
       $...$
       
        $\mathbf{h}_i, \mathbf{o}_i, \alpha_i \gets \text{Gaussian } i \text{ parameters}$\;
        
        \tcp{Ray-Ellipsoid Intersection}
        $\text{rayEllipsoidIntersection}~\myeqref{equ:rayEllipsoidIntersection}$ \; 
        
        \tcp{Use Midpoint for Depth}
        $d_i \gets \text{obtainDepth} ~\myeqref{eq:midpoint}$ \;
        \tcp{Alpha-Blending for All Attributes}
        \For{each $ch$ in $C_h$}{
        $\mathbf{C}_{[ch]} \gets \mathbf{C} + \mathbf{h}_i \cdot \alpha_i \cdot T$\;
        }
        \For{each $ch$ in $C_o$}{
        $\mathbf{O}_{[ch]} \gets \mathbf{O} + \mathbf{o}_i \cdot \alpha_i \cdot T$\;
        }
        $D \gets D + d_i \cdot \alpha_i \cdot T$\;
        $K \gets K + t_i \cdot \alpha_i \cdot T$\;
        $...$
    }
    \tcp{Pixel-Wise Assignment}
    $\tilde{\mathbf{C}}[x,y] \gets \mathbf{C}$;\quad
    $\tilde{\mathbf{O}}[x,y] \gets \mathbf{O}$\;
    $\tilde{\mathbf{D}}[x,y] \gets D$;\quad
    $\tilde{\mathbf{K}}[x,y] \gets K$\;
    \text{Save} $T_\text{final}=T$\;
}
\end{algorithm}

\noindent\textbf{Semantic Rendering and Backpropagation}.
\label{sec:semantic_rendering}
We render a 2D semantic logits $\tilde{\mathbf{O}} \in \mathbb{R}^{H \times W \times C_{\mathbf{o}}}$ along with rasterization pipeline. The forward process for the semantic logits in pixel $p=(u,v)$ and channel $ch$ is defined as
\begin{equation}
\tilde{\mathbf{O}}_{[ch]}(u,v) = \sum_{i\in \mathcal{N}}^{} \alpha_i  \mathbf{o}_{i[ch]}\prod_{j=1}^{i-1} (1 - \alpha_j).
\label{eq:semantic_render}
\end{equation}
Then the rendered semantic logits $\tilde{\mathbf{O}}$ will be compared to its ground truth $\mathbf{O}^{\text{gt}}$ using Cross-Entropy Loss, resulting in semantic loss $\mathcal{L}_{\text{seg}}$.
Let $\frac{\partial \mathcal{L}_{\text{seg}}}{\partial \tilde{\mathbf{o}}_{[ch]}}$ be the gradient of the loss with respect to the rendered $ch$-th channel of a specific pixel.
The chain rule for the gradient of $\mathcal{L}_{\text{seg}}$ with respect to a specific semantic channel $\mathbf{o}_{i[ch]}$ of a Gaussian $\mathcal{G}$ is given by
\begin{equation}
\label{eq:gradiento}
\frac{\partial \mathcal{L}_{\text{seg}}}{\partial \mathbf{o}_{i[ch]}} = \frac{\partial \mathcal{L}_{\text{seg}}}{\partial \tilde{\mathbf{O}}_{[ch]}} \cdot \frac{\partial \tilde{\mathbf{O}}_{[ch]}} {\partial \mathbf{o}_{i[ch]}},
\end{equation}
where $\frac{\partial \tilde{\mathbf{O}}_{[ch]}} {\partial \mathbf{o}_{i[ch]}}$ can be derived from~\myeqref{eq:semantic_render}.


\begin{table*}[!htb]
\centering
\caption{Quantitative comparison experiments on the Replica dataset show that our method significantly outperforms other baselines.
}
\resizebox{\linewidth}{!}{
\begin{tabular}{clcccccccccccc}
\hline
\multicolumn{2}{c}{Dataset}          & \multicolumn{3}{c}{O0S1}                                                                         & \multicolumn{3}{c}{O1S2}                                                                         & \multicolumn{3}{c}{R1S2}                                                                         & \multicolumn{3}{c}{R2S2}                                                                         \\
\multicolumn{2}{c}{Method $|$ Metrics} & Abs.Rel ↓                        & RMSE↓                           & CS ↑                             & Abs.Rel↓                        & RMSE↓                           & CS ↑                            & Abs.Rel↓                        & RMSE↓                           & CS ↑                            & Abs.Rel↓                        & RMSE ↓                          & CS↑                             \\ \hline
\multicolumn{2}{c}{GeoGaussian~\cite{li2024geogaussian}}      & 0.5828                         & 0.8006                         & 0.3046                         & 2.8959                         & 1.1092                         & 0.1861                         & 0.1336                         & 0.3236                         & 0.5700                         & 0.1082                         & 0.2592                         & 0.5998                         \\
\multicolumn{2}{c}{3DGS~\cite{3dgs}}             & 1.2231                         & 0.9183                         & 0.2304                         & 0.8711                         & 0.6238                         & 0.3201                         & 0.2134                         & 0.4092                         & 0.4963                         & 1.4422                         & 1.0972                         & 0.2905                         \\
\multicolumn{2}{c}{Scaffold-GS~\cite{scaffoldgs}}      & 2.2857                         & 1.2102                         & 0.2195                         & 1.4609                         & 0.9413                         & 0.2854                         & 4.0575                         & 1.4786                         & 0.1555                         & -                              & -                              & 0.1286                         \\
\multicolumn{2}{c}{2DGS~\cite{huang20242d}}             & \cellcolor[HTML]{FFD9B3}0.0089 & \cellcolor[HTML]{FFD9B3}0.0679 & \cellcolor[HTML]{FFF5B3}0.8775 & \cellcolor[HTML]{FFD9B3}0.0172 & \cellcolor[HTML]{FFD9B3}0.0399 & \cellcolor[HTML]{FFF5B3}0.9299 & \cellcolor[HTML]{FFD9B3}0.0451 & \cellcolor[HTML]{FFD9B3}0.1400 & \cellcolor[HTML]{FFD9B3}0.9112 & \cellcolor[HTML]{FFD9B3}0.0735 & \cellcolor[HTML]{FFD9B3}0.1759 & \cellcolor[HTML]{FFD9B3}0.8480 \\
\multicolumn{2}{c}{Normal-GS~\cite{normalgs}}        & \cellcolor[HTML]{FFF5B3}0.0074 & \cellcolor[HTML]{FFF5B3}0.0657 & \cellcolor[HTML]{FFD9B3}0.8755 & \cellcolor[HTML]{FFF5B3}0.0154 & \cellcolor[HTML]{FFF5B3}0.0344 & \cellcolor[HTML]{FFD9B3}0.9234 & \cellcolor[HTML]{FFF5B3}0.0243 & \cellcolor[HTML]{FFF5B3}0.0766 & \cellcolor[HTML]{FFF5B3}0.9234 & \cellcolor[HTML]{FFF5B3}0.0192 & \cellcolor[HTML]{FFF5B3}0.0589 & \cellcolor[HTML]{FFF5B3}0.8987 \\ \hline
\multicolumn{2}{c}{Ours}             & \cellcolor[HTML]{C0E2CA}0.0062 & \cellcolor[HTML]{C0E2CA}0.0310 & \cellcolor[HTML]{C0E2CA}0.8880 & \cellcolor[HTML]{C0E2CA}0.0015 & \cellcolor[HTML]{C0E2CA}0.0073 & \cellcolor[HTML]{C0E2CA}0.9686 & \cellcolor[HTML]{C0E2CA}0.0136 & \cellcolor[HTML]{C0E2CA}0.0467 & \cellcolor[HTML]{C0E2CA}0.9545 & \cellcolor[HTML]{C0E2CA}0.0144 & \cellcolor[HTML]{C0E2CA}0.0465 & \cellcolor[HTML]{C0E2CA}0.9076 \\ \hline
\end{tabular}}
\label{tab:depthnormal_rendering}
\end{table*}

\begin{figure*}[tbp]
\centering
\includegraphics[width=\linewidth]{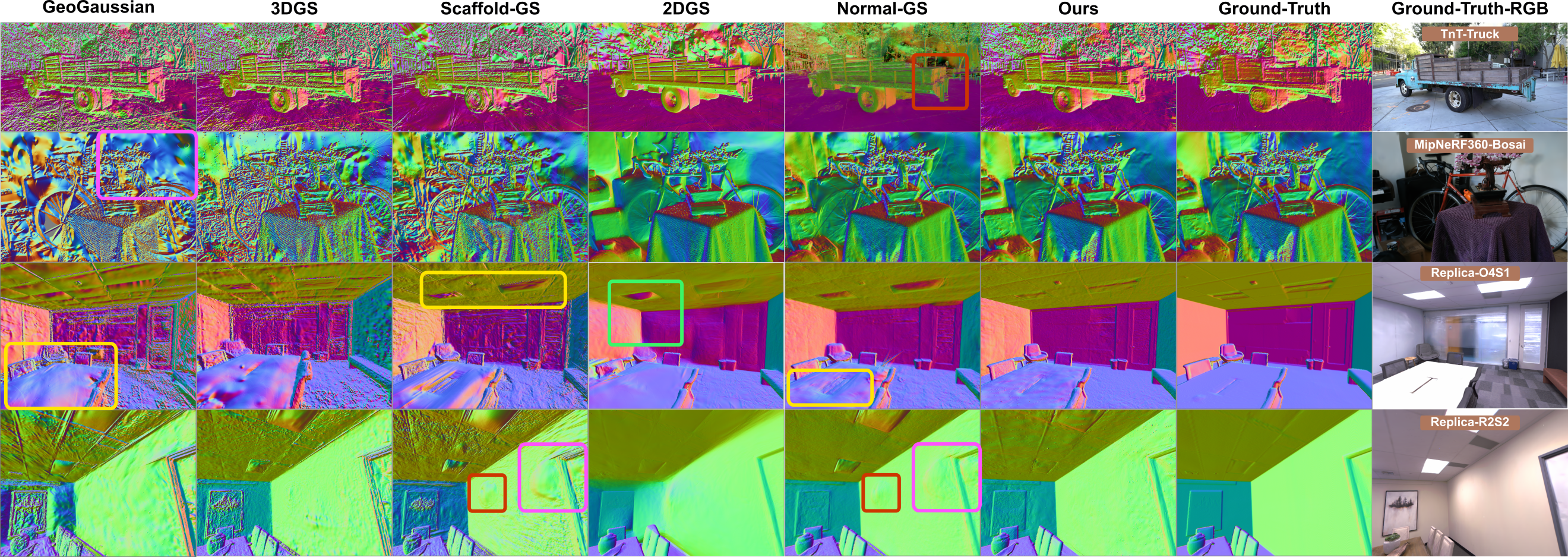}
\caption{Across various dataset tests, our method demonstrates impressive results in geometric reconstruction. Even in textureless areas and edge regions, it remains capable of recovering fine geometric details.}
\label{fig:depthnormal_rendering}
\vspace{-1em}
\end{figure*}


\noindent\textbf{Tile-based Forward Rasterization}. 
We extend the original 3DGS's tile-based rendering pipeline~\cite{3dgs}. During rasterization, for each pixel $p=(u,v)$, we calculate the direction $\mathbf{d}$ of the camera ray $r(t)$ and then compute the intersection $t_1/t_2$ with the Gaussian projected onto this pixel. Finally, we obtain the corresponding attributes using alpha blending rendering. The multimodal forward rasterization pipeline is outlined in~\myalgref{alg:forward_render}. The corresponding backpropagation pipeline is described in the supplementary materials.

\noindent\textbf{Loss Function}.
For the RGB image loss, we employ a combination of L1 loss \( \mathcal{L}_{l1} \) and SSIM~\cite{wang2004image} loss \( \mathcal{L}_{\text{ssim}} \). For the normal loss \( \mathcal{L}_{\text{normal}} \), we use cosine similarity, which is calculated as
\begin{equation}
\mathcal{L}_{normal} = 1-\text{Cosine Similarity}(\tilde{\mathbf{N}},\,\tilde{\mathbf{N}}{_\text{gt}}).
\label{eq:csloss}
\end{equation}
Incorporating the gradient factor loss \( \mathcal{L}_K \) and semantic loss \( \mathcal{L}_{seg} \), we construct the comprehensive loss function as
\begin{align}
\mathcal{L}_{all} = & \, \lambda_1 \mathcal{L}_{l1} + \lambda_2 \frac{|\mathcal{L}_{l1}|\mathcal{L}_{ssim}}{|\mathcal{L}_{ssim}|} + \lambda_3 \frac{|\mathcal{L}_{l1}|\mathcal{L}_{normal}}{|\mathcal{L}_{normal}|} \notag \\
& \lambda_4 \frac{|\mathcal{L}_{l1}|\mathcal{L}_{depth}}{|\mathcal{L}_{depth}|} + \lambda_5 \frac{|\mathcal{L}_{l1}|\mathcal{L}_{seg}}{|\mathcal{L}_{seg}|} + \lambda_6 \frac{|\mathcal{L}_{l1}|\mathcal{L}_{K}}{|\mathcal{L}_{K}|}.
\end{align}
where weights are applied as $\lambda_1=1,\lambda_2=\lambda_3=\lambda_4=\lambda_4=\lambda_5=\lambda_6=0.1$.

\noindent\textbf{Implementation Details.} Our framework is implemented by CUDA using the LibTorch framework~\cite{libtorch}, incorporating CUDA code for Gaussian Splatting and trained on a server with a 2.60\,GHz Intel(R) Xeon(R) Platinum 8358P CPU, 1T\,GB RAM, and a NVIDIA RTX 4090 24\,GB GPU. In all scequences, the hyprparameters used in our experiments are listed in the supplementary materials.

%% file: sec/4_exps.tex
\mysection{Experiments}
In this section, we first introduce the experimental setup in \mysecref{sec:expsetup}, including datasets, baselines, and parameter settings, etc. Given that RGB rendering is not the central focus of our study, the corresponding comparison will be provided in the supplementary material accordingly. 
Then in
\mysecref{sec:geo_rendering}, we evaluate the geometry rendering performance (e.g. depth and normal). In \mysecref{sec:seg_rendering}, we evaluate the semantic rendering performance. In \mysecref{sec:time-memory}, we provide the analysis of running time and GPU memory consumption of the framework. \mysecref{sec:abla} shows ablation experiments of the proposed framework.

\mysubsection{Experiment Setup}
\label{sec:expsetup}

\noindent\textbf{Datasets.}
We conduct comparative experiments on widely used datasets, including 7 scenes from Mip-NeRF~\cite{barron2022mip}, 2 scenes from Tanks \& Temples~\cite{knapitsch2017tanks} and 2 scenes from Deep Blending~\cite{hedman2018deep}. All of these datasets do not provide precise geometric and semantic ground truth. To address this, we use the results obtained from the depth rendering pipeline in Normal-GS~\cite{normalgs} as geometric priors for supervision. At the same time, we utilize open vocabulary queries to pre-process semantic labels with OpenSeed~\cite{zhang2023simple}. For accurate quantitative comparison experiments, we use the Replica~\cite{straub2019replica} simulation dataset, which provides complete multimodal ground truth. We employ the pre-rendered Replica dataset from HiSplat~\cite{tang2024hisplat}, which includes 14 motion sequences across 7 scenes. The initial points in the Replica~\cite{straub2019replica} dataset are obtained through mesh sampling, with a sampling interval of 5 in all sequences.

\noindent\textbf{Baselines and Metrics.}
We compare our method with the existing SOTA NeRF-based dense framework Mip-NeRF~\cite{barron2022mip}, iNGP~\cite{muller2022instant} and 3DGS framework~\cite{3dgs}, Scaffold-GS~\cite{scaffoldgs}, GeoGaussian~\cite{li2024geogaussian}, 2DGS~\cite{huang20242d}, Normal-GS~\cite{normalgs}, SemanticGaussian~\cite{guo2024semantic}. We evaluate the RGB rendering performance using PSNR, SSIM~\cite{wang2004image}, and LPIPS~\cite{zhang2018unreasonable}. The evaluation metrics for depth estimation we used are \textit{Abs.Rel}. and RMSE. The evaluation metric we used for normal estimation is cosine similarity (CosSimi), for semantic segmentation, we employed mean mIoU.

\mysubsection{Geometry Rendering Evaluation}
\label{sec:geo_rendering}
We conduct detailed qualitative and quantitative comparisons of depth and normal estimation against methods~\cite{3dgs}~\cite{scaffoldgs}~\cite{jiang2024gaussianshader}~\cite{li2024geogaussian}~\cite{normalgs}~\cite{huang20242d} on the Replica~\cite{straub2019replica} dataset. The comparison results are shown in~\myfigref{fig:depthnormal_rendering} and~\mytabref{tab:depthnormal_rendering}. Normal-GS~\cite{normalgs} and our method optimize the surface normals, and both qualitatively and quantitatively outperform other 3DGS frameworks. Our approach also exhibits superior geometric accuracy in some extreme scenarios, particularly in texture-less areas, surpassing even Normal-GS~\cite{normalgs} and 2DGS~\cite{huang20242d}. More comparisons about geometry rendering can be found in supplementary materials.

\begin{figure}[t]
\centering
\includegraphics[width=\linewidth]{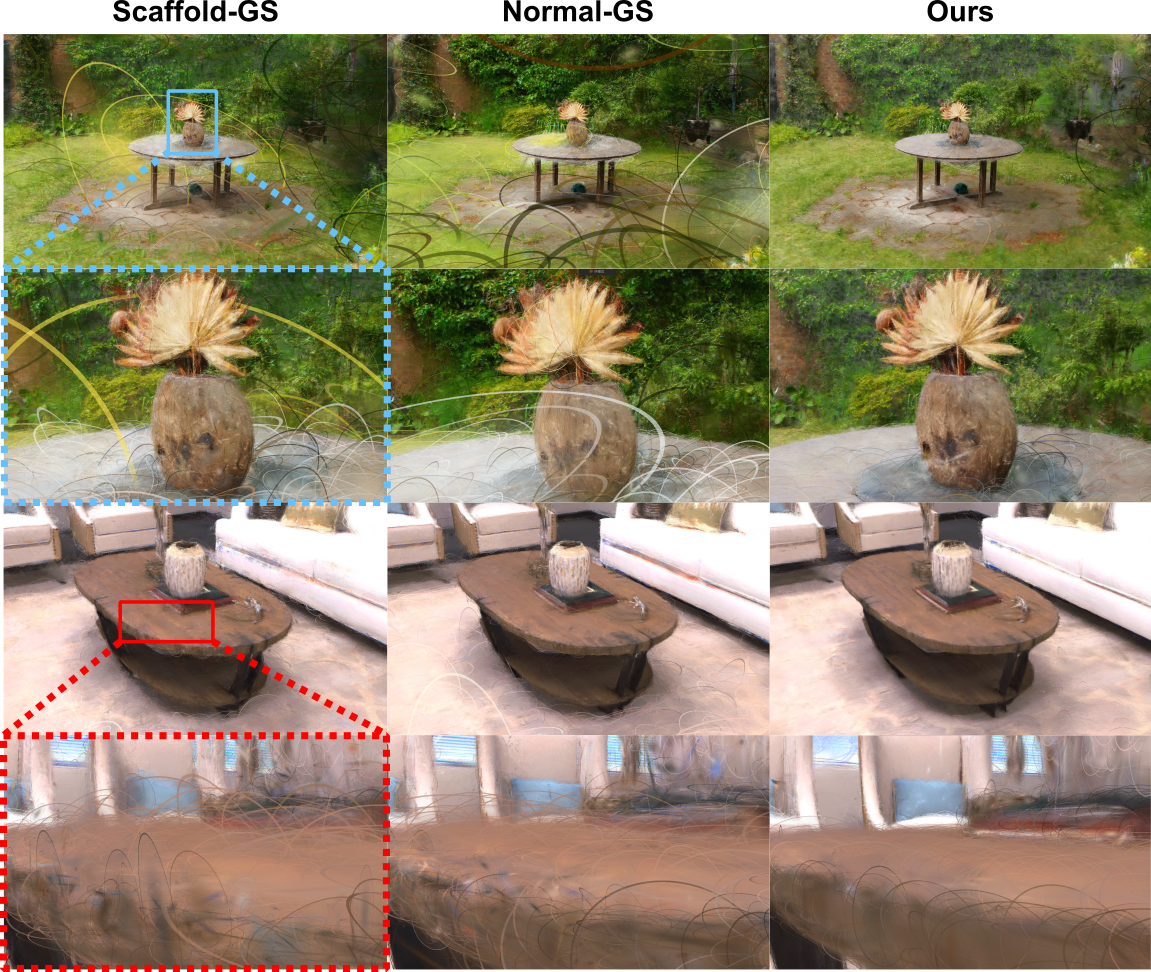}
\caption{A qualitative demonstration showing that the Gaussian primitives reconstructed by our method are geometrically closer to the real surface.}
\label{fig:fit3d}
\vspace{-1em}
\end{figure}

\begin{figure}[t]
\centering
\includegraphics[width=\linewidth]{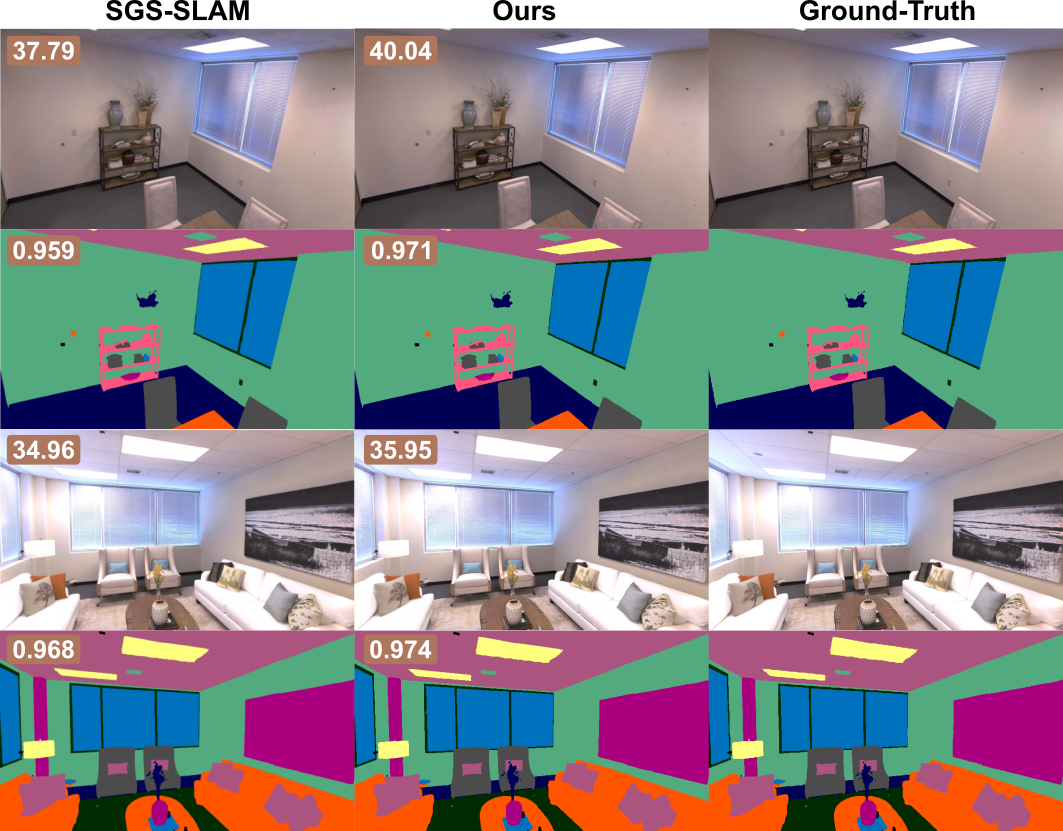}
\caption{Quantitative comparison between our method and SGS-SLAM~\cite{sgsslam} in terms of RGB rendering and semantic segmentation clearly shows that our approach outperforms SGS-SLAM. The top-left corner of the image displays the PSNR value computed against the ground truth image.}
\label{fig:seg_rendering}
\vspace{-1em}
\end{figure}

\begin{table}[!htb]
\centering
\caption{Quantitative geometric comparison between our method and other methods after sampling the reconstructed Gaussian model, compared to the initial points or mesh points.
}
\resizebox{\linewidth}{!}{
\begin{tabular}{clcccccccc}
\hline
\multicolumn{2}{c}{Dataset}          & \multicolumn{4}{c}{MipNeRF-Garden}                                                                                               & \multicolumn{4}{c}{Replica-Room0}                                                                                                \\
\multicolumn{2}{c}{Method $|$ Metrics} & Radio↓                          & Mean↓                           & Std↓                            & Hau-Dis↓                       & Radio   ↓                       & Mean↓                           & Std ↓                           & Hau-Dis ↓                      \\ \hline
\multicolumn{2}{c}{Scaffold-GS~\cite{scaffoldgs}}      & \cellcolor[HTML]{FFD9B3}0.0390 & \cellcolor[HTML]{FFF5B3}0.0299 & \cellcolor[HTML]{FFD9B3}0.0383 & \cellcolor[HTML]{FFF5B3}0.592 & \cellcolor[HTML]{FFF5B3}0.0476 & \cellcolor[HTML]{FFD9B3}0.0120 & \cellcolor[HTML]{FFD9B3}0.0083 & \cellcolor[HTML]{FFD9B3}0.141 \\
\multicolumn{2}{c}{2DGS~\cite{huang20242d}}             & 0.0607                         & 0.0984                         & 0.1272                         & 1.091                         & 0.0628                         & 0.0519                         & 0.0244                         & 0.462                         \\
\multicolumn{2}{c}{Normal-GS~\cite{normalgs}}        & \cellcolor[HTML]{FFF5B3}0.0388 & \cellcolor[HTML]{FFD9B3}0.0303 & \cellcolor[HTML]{FFF5B3}0.0377 & \cellcolor[HTML]{FFD9B3}0.936 & \cellcolor[HTML]{FFD9B3}0.0477 & \cellcolor[HTML]{FFF5B3}0.0120 & \cellcolor[HTML]{FFF5B3}0.0082 & \cellcolor[HTML]{FFF5B3}0.131 \\
\multicolumn{2}{c}{Ours}             & \cellcolor[HTML]{C0E2CA}0.0341 & \cellcolor[HTML]{C0E2CA}0.0272 & \cellcolor[HTML]{C0E2CA}0.0298 & \cellcolor[HTML]{C0E2CA}0.405 & \cellcolor[HTML]{C0E2CA}0.0339 & \cellcolor[HTML]{C0E2CA}0.0137 & \cellcolor[HTML]{C0E2CA}0.0081 & \cellcolor[HTML]{C0E2CA}0.078 \\ \hline
\end{tabular}}
\label{tab:fit3d}
\end{table}

We further compare the geometric impact after modifying our depth rendering method. Firstly, we complete the reconstruction process of a 3D Gaussian scene and then identify a detailed area in the reconstructed scene (as shown in ~\myfigref{fig:fit3d}). It is evident that our method, compared to others, results in optimized Gaussian primitives that tend to be closer to the actual surface of the object. This is because our method integrates rotation and scale attribute into the optimization process, thereby achieving geometric consistency. Additionally, we perform point sampling~\cite{li2024geogaussian} of the Gaussian primitives within specific small regions, followed by a Z-score statistical comparison with the initial ground truth and a Hausdorff distance comparison, as shown in~\mytabref{tab:fit3d}. It is evident that our method achieves consistent geometric surface reconstruction. The calculation method using the Z-score statistic is explained in the supplementary materials.

\begin{table}[htb!]
\centering
\caption{Quantitative comparison with the SGS-SLAM~\cite{sgsslam} method in terms of RGB rendering, semantic rendering and FPS. TT represents training time (minutes). Best results are underlined.
}
\resizebox{0.99\linewidth}{!}{
\begin{tabular}{@{}clcccccccccc@{}}
\toprule
\multicolumn{2}{c}{Dataset}          & All        & \multicolumn{3}{c}{Room0}               & \multicolumn{3}{c}{Room2}               & \multicolumn{3}{c}{Office2}             \\
\multicolumn{2}{c}{Method $|$ Metrics} & TT↓         & PSNR ↑       & FPS ↑        & mIoU   ↑     & PSNR  ↑      & FPS     ↑    & mIoU↑        & PSNR   ↑     & FPS   ↑      & mIoU   ↑     \\ \midrule
\multicolumn{2}{c}{SGS-SLAM~\cite{sgsslam}}         & 212.6      & 36.28       & 96.4        & 0.976       & 39.14       & 72.2        & 0.978       & 36.94       & 112.3       & 0.944       \\
\multicolumn{2}{c}{Ours}             & {\ul 32.2} & {\ul 36.66} & {\ul 184.2} & {\ul 0.987} & {\ul 40.06} & {\ul 234.4} & {\ul 0.984} & {\ul 38.16} & {\ul 229.3} & {\ul 0.975} \\ \bottomrule
\end{tabular}}
\label{tab:seg_rendering}
\end{table}

\mysubsection{Semantic Rendering Evalution}
\label{sec:seg_rendering}

We compare our method with the SGS-SLAM~\cite{sgsslam} framework, which employs a task-specific Gaussian map representation and is supervised using semantic result images for semantic tasks, differing from our unified framework that predicts semantic logits. The comparison across multiple scenes is shown in~\myfigref{fig:seg_rendering} and~\mytabref{tab:seg_rendering}.
From the quantitative results, it can be observed that our method achieves superior performance in both image rendering quality (PSNR) and semantic segmentation quality (mIoU) within a very short training time. Thanks to our unified rendering architecture, the image rendering FPS is also fast. Additional results can be found in our supplementary video.

\begin{figure}[t]
\centering
\includegraphics[width=\linewidth]{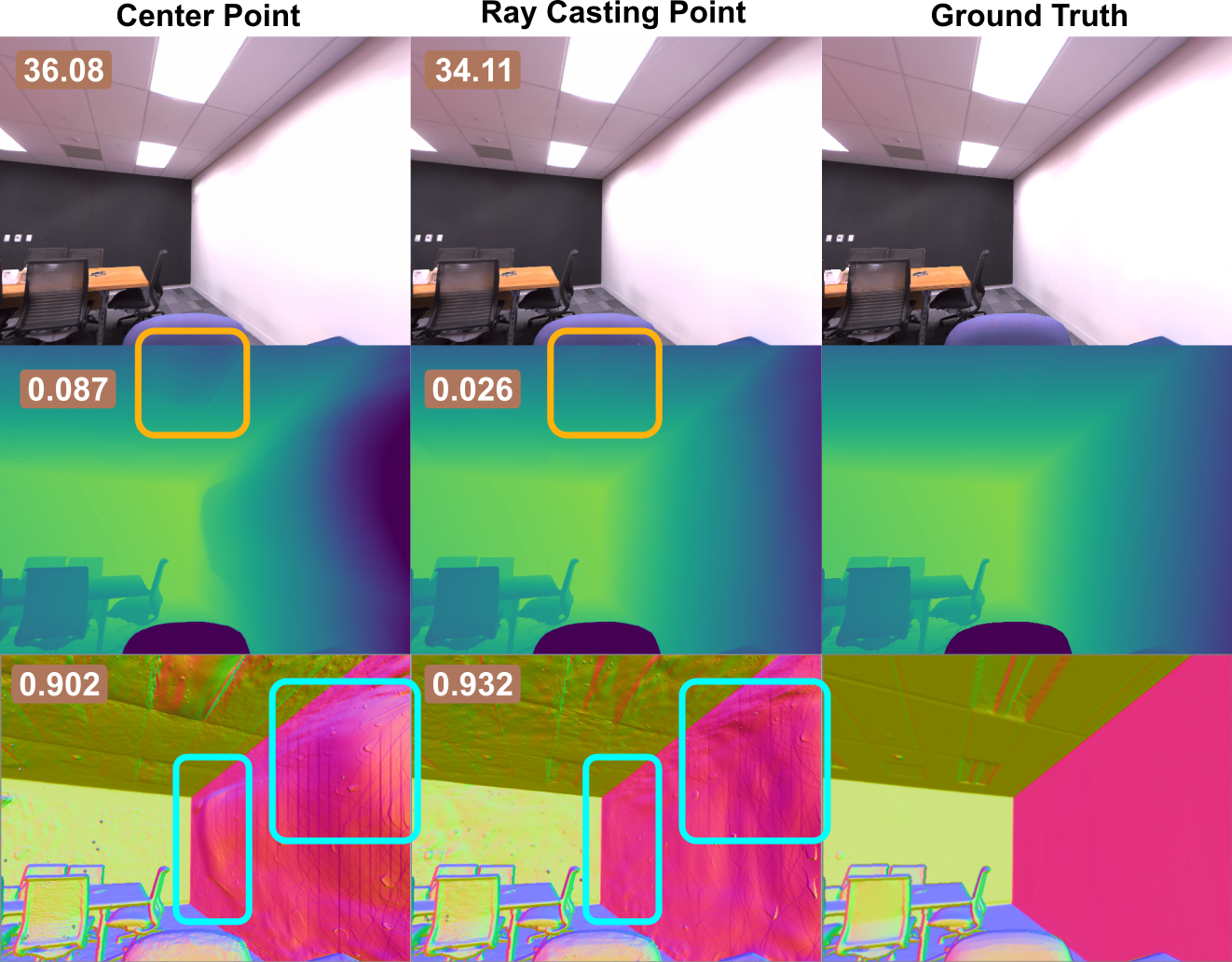}
\caption{The use of a ray casting-based depth rendering approach significantly enhances the algorithm's robustness in textureless regions. Furthermore, the Gaussian splats provide a tighter fit to the actual object surfaces.}
\label{fig:normal_abla}
\vspace{-1em}
\end{figure}

\begin{table}[ht]
\caption{Quantitative comparison results in terms of Gaussian splat count and rendering FPS.}
\centering
\label{tab:mem_time}
\resizebox{\linewidth}{!}{
\begin{tabular}{@{}clcccccccc@{}}
\toprule
\multicolumn{2}{c}{Dataset}                            & \multicolumn{2}{c}{Mip-NeRF360}                                & \multicolumn{2}{c}{Tanks \& Temples}                             & \multicolumn{2}{c}{Deep Blending}                              & \multicolumn{2}{c}{Replica}                                    \\
\multicolumn{2}{c}{Method $|$ Metrics}                   & Count                         ↓ & FPS ↑                          & Count                         ↓ & FPS ↑                     & Count                         ↓ & FPS ↑ & Count                         ↓ & FPS ↑                         \\ \midrule
\multicolumn{2}{c}{{\color[HTML]{9B9B9B} GeoGaussian~\cite{li2024geogaussian}}} & {\color[HTML]{9B9B9B} 126021}  & {\color[HTML]{9B9B9B} 440.9}  & {\color[HTML]{9B9B9B} 94755}   & {\color[HTML]{9B9B9B} 441.1}  & {\color[HTML]{9B9B9B} 53457}   & {\color[HTML]{9B9B9B} 436.2}  & {\color[HTML]{9B9B9B} 100877}  & {\color[HTML]{9B9B9B} 423.6}  \\
\multicolumn{2}{c}{3DGS~\cite{3dgs}}                               & 504167                         & 128.7                         & 421167                         & 136.9                         & 285394                         & 162.4                         & 543581                         & 137.5                         \\
\multicolumn{2}{c}{Scaffold-GS~\cite{scaffoldgs}}                        & \cellcolor[HTML]{FFF5B3}301839 & \cellcolor[HTML]{C0E2CA}240.1 & \cellcolor[HTML]{FFD9B3}247081 & \cellcolor[HTML]{C0E2CA}249.1 & \cellcolor[HTML]{FFF5B3}182304 & \cellcolor[HTML]{C0E2CA}324.0 & \cellcolor[HTML]{FFD9B3}235446 & \cellcolor[HTML]{C0E2CA}346.2 \\
\multicolumn{2}{c}{Normal-GS~\cite{normalgs}}                          & \cellcolor[HTML]{FFD9B3}318064 & \cellcolor[HTML]{FFF5B3}180.3 & \cellcolor[HTML]{FFF5B3}226283 & \cellcolor[HTML]{FFF5B3}178.7 & \cellcolor[HTML]{FFD9B3}185243 & \cellcolor[HTML]{FFD9B3}250.7 & \cellcolor[HTML]{FFF5B3}204358 & \cellcolor[HTML]{FFD9B3}180.0 \\
\multicolumn{2}{c}{Ours}                               & \cellcolor[HTML]{C0E2CA}161081 & \cellcolor[HTML]{FFD9B3}169.8 & \cellcolor[HTML]{C0E2CA}119947 & \cellcolor[HTML]{FFD9B3}163.1 & \cellcolor[HTML]{C0E2CA}48970  & \cellcolor[HTML]{FFF5B3}296.7 & \cellcolor[HTML]{C0E2CA}169273 & \cellcolor[HTML]{FFF5B3}197.8 \\ \bottomrule
\end{tabular}
}
\vspace{-1em}
\end{table}

\mysubsection{Runtime and Memory Cost Analysis}
\label{sec:time-memory}

In~\mytabref{tab:mem_time}, we present a comparison of the Count of Gaussians and rendering FPS against other methods. Our approach employs an effective differentiable pruning module, which removes a significant number of redundant Gaussians. As a result, the number of Gaussian splats in our reconstructed scenes is the smallest. However, since our method requires rendering multiple modalities simultaneously, the FPS experiences a certain degree of decrease. Nevertheless, this does not severely hinder our rendering speed. In some scenarios, our rendering speed remains superior even compared to Normal-GS~\cite{normalgs}.

\mysubsection{Ablation Study}
\label{sec:abla}
\noindent\textbf{Geometry-Aware Rasterization.}
We compare the performance of depth rasterization that considers geometric information with that of depth rasterization using only the center points in terms of image quality. From the qualitative results presented in~\myfigref{fig:normal_abla}, it can be observed that incorporating set information forces the Gaussian spheres to align more closely with the actual surface of the objects. More quantitative ablation experiments on each modules can be found in our supplementary materials.

\begin{figure}[t]
\centering
\includegraphics[width=\linewidth]{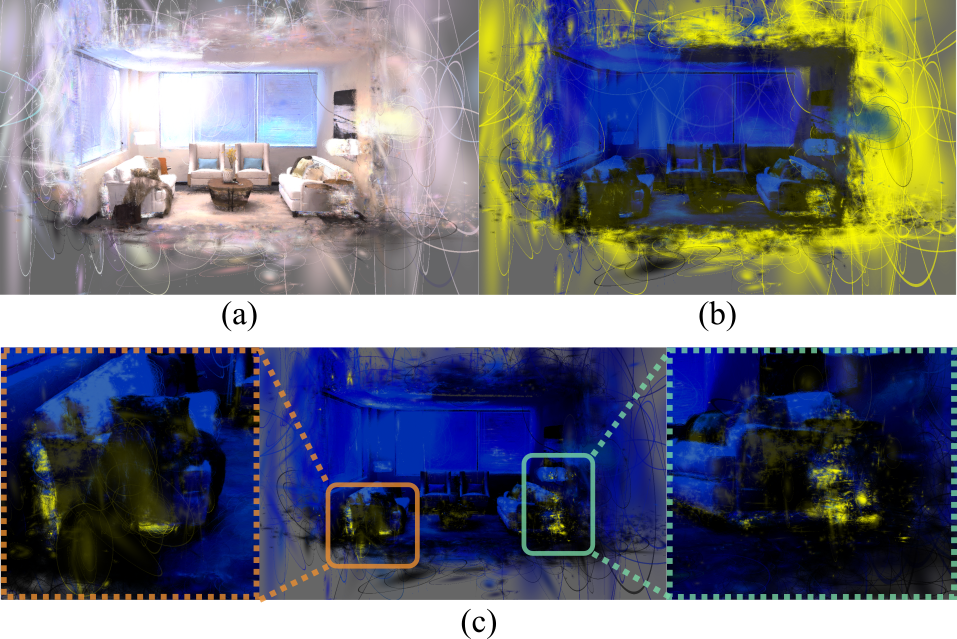}
\caption{\textbf{Visualization of the distribution of the gradient factor.} To enhance the visualization of the gradient factor, the brightness of all Gaussians is uniformly minimized, resulting in blue regions as rendered by SuperSplat~\cite{supersplat}. (a) Scene with original brightness. (b) Areas highlighted in yellow indicate regions where the gradient factor $k > 1.5$. (c) Areas highlighted in yellow indicate regions where $k < 0.5$.}
\label{fig:prune_vis}
\vspace{-1em}
\end{figure}

\noindent\textbf{Differentiable Pruning.}
We conduct continuous optimization of the gradient factor without pruning any Gaussians to examine its distribution, as shown in~\myfigref{fig:prune_vis}. It can be observed that regions with anomalous gradient factor, which deviate significantly from the ideal value of 1, are typically areas lacking observational data or corresponding to occluded edges. In these regions, due to insufficient supervisory information from images, the gradients fail to converge to the desired value. Such areas represent environmental redundancy that can be removed to improve computational efficiency.

%% file: sec/6_conclusion.tex
\mysection{Conclusion}
In this paper, we present \textbf{UniGS}, a \textit{unified} and differentiable framework for high-fidelity multimodal 3D reconstruction using 3DGS. Our rasterization pipeline effectively renders photo-realistic RGB images, accurate depth maps, consistent surface normals, and semantic labels simultaneously. By implementing differentiable ray-ellipsoid intersection, we optimize rotation and scaling parameters with CUDA-accelerated analytic gradient backpropagation, ensuring geometric consistency in surface. We also introduce a learnable attribute for efficient pruning of minimally contributing Gaussians. Extensive experiments demonstrate state-of-the-art performance in terms of both reconstruction accuracy across all modalities and storage efficiency.

%% file: sec/X_suppl.tex
\clearpage
\setcounter{page}{1}
\maketitlesupplementary
\externaldocument{sec/3_method}
\section{Methodology}

In~\mysecref{sec:depth_render}, we provide additional details on solving the intersection points between pixel rays and Gaussians. In ~\mysecref{sec:depth_backward}, we focuse on deriving the analytical solution for propagating depth gradients to the Gaussian parameters, specifically addressing the transition from depth to rotation, scale and positions. In~\mysecref{sec:normal_render}, we include the implementation details that are omitted in~\myalgref{alg:normal_estimation}. In~\mysecref{sec:normal_backward}, we develop the derivation of the gradient propagation from the cosine similarity loss of the normal estimation to the rendered depth map, enabling the normal gradients to flow back to depth gradient. This connection allows the normal gradients to be backpropagated further, thus linking the normal consistency loss directly to the Gaussian parameters. In~\mysecref{sec:diff_prune}, we expand the discussion on differentiable pruning, with a particular focus on elucidating why $k=1$ serves as the optimization objective. Finally, in~\mysecref{sec:backward_render}, we provide the implementation process for the backward rasterization gradient propagation corresponding to~\myalgref{alg:forward_render}.

\subsection{Ray-Eclipse Intersection Problem}
\label{sec:depth_render}
The ray-eclipse intersection problem in~\myeqref{equ:rayEllipsoidIntersection} aim to solve for $t$ in the equation of a unit sphere
\begin{equation}
\| \mathbf{v}_{s} + t \mathbf{d}_{s} \|^2 = 1,
\end{equation}
which simplifies to the quadratic equation:
\begin{equation}
at^2 + bt + c = 0,
\end{equation}
where the coefficients are defined as:
\begin{equation}
\label{eq:abccoeff}
\begin{aligned}
a &= \mathbf{d}_{s} \cdot \mathbf{d}_{s}, \\
b &= 2 \mathbf{v}_{s} \cdot \mathbf{d}_{s}, \\
c &= \mathbf{v}_{s} \cdot \mathbf{v}_{s} - 1.
\end{aligned}
\end{equation}
The solutions are given by $t = \frac{-b \pm \sqrt{\Delta}}{2a}$, where $\Delta = b^2 - 4ac$. This is a simple quadratic equation solving problem.

\subsection{Backward Propagation for Depth Rendering}\label{sec:depth_backward}
The gradient of the loss $\mathcal{L}_{depth}$ with respect to the a single Gaussian parameters $\boldsymbol{\theta} \in \{\boldsymbol{\mu}, \mathbf{s}, \mathbf{R}\}$ is propagated by
\begin{equation}
\frac{\partial \mathcal{L}_{depth}}{\partial \boldsymbol{\theta}} = \frac{\partial \mathcal{L}_{depth}}{\partial \tilde{\mathbf{D}}} \cdot \frac{\partial \tilde{\mathbf{D}}}{\partial d_i}\cdot \frac{\partial d_i}{\partial \boldsymbol{\theta}},
\end{equation}
where $\frac{\partial \tilde{\mathbf{D}}}{\partial d_i}$ can be calculated from~\myeqref{equ:depth_render}.
The primary challenge is to compute the gradients of $d_i$ with respect to the Gaussian parameters $\boldsymbol{\mu}$, $\mathbf{s}$, and $\mathbf{R}$, which requires backpropagating through the ray-ellipsoid intersection. The gradient flow is structured as
\begin{equation}
\begin{aligned} 
\frac{\partial d_i}{\partial \boldsymbol{\theta}} & = \frac{\partial d_i}{\partial t_{\text{mid}}} \cdot \frac{\partial t_{\text{mid}}}{\partial (a, b, c)} \cdot \frac{\partial (a, b, c)}{\partial (\mathbf{v}_{s}, \mathbf{d}_{s})} \\& \cdot \frac{\partial (\mathbf{v}_{s}, \mathbf{d}_{s})}{\partial (\mathbf{v}_{l}, \mathbf{d}_{l})} \cdot \frac{\partial (\mathbf{v}_{l}, \mathbf{d}_{l})}{\partial \boldsymbol{\theta}}. 
\end{aligned}
\end{equation}
The key steps in this chain will be derived below.

\noindent\textbf{Gradient of Depth w.r.t. $t_{\text{mid}}$}.
The depth $d_i$ of the intersection point is a function of the midpoint ray parameter $t_{\text{mid}}$. The gradient is computed as the dot product of the ray direction $\mathbf{d}$ with the $z$-axis of the camera view matrix $\mathbf{R}_{\text{world}}^{\text{cam}}$:
\begin{align}
\label{eq:d->tmid}
\frac{\partial d_i}{\partial t_{\text{mid}}} = &\ \mathbf{R}_{\text{world}}^{\text{cam}}[2,0] \cdot \mathbf{d}\to x + \mathbf{R}_{\text{world}}^{\text{cam}}[2,1] \cdot \mathbf{d}\to y \notag \\
&\ + \mathbf{R}_{\text{world}}^{\text{cam}}[2,2] \cdot \mathbf{d}\to z.
\end{align}

\noindent\textbf{Gradient of $t_{\text{mid}}$ w.r.t. Quadratic Coefficients.}
The midpoint parameter $t_{\text{mid}} = (t_1 + t_2)/2 = -b / (2a)$ is a function of the quadratic coefficients $a$ and $b$. Its derivatives are
\begin{equation}
\label{eq:tmid->abc}
\begin{aligned}
\frac{\partial t_{\text{mid}}}{\partial a} = \frac{b}{2a^2},\,
\frac{\partial t_{\text{mid}}}{\partial b} = -\frac{1}{2a},\,
\frac{\partial t_{\text{mid}}}{\partial c} = 0.
\end{aligned}
\end{equation}

\noindent\textbf{Gradients of Coefficients w.r.t. Local Coordinates}.
The coefficients $a$, $b$, and $c$ are functions of the scaled local ray origin $\mathbf{v}_{s}$ and direction $\mathbf{d}_{s}$ from~\myeqref{eq:abccoeff}.
Their gradients are
\begin{equation}
\label{eq:abc->vs}
\begin{aligned}
\frac{\partial a}{\partial \mathbf{d}_{s}} &= 2 \mathbf{d}_{s}, \,
\frac{\partial b}{\partial \mathbf{v}_{s}} = 2 \mathbf{d}_{s},\\ \frac{\partial b}{\partial \mathbf{d}_{s}} &= 2 \mathbf{v}_{s}, \,
\frac{\partial c}{\partial \mathbf{v}_{s}} = 2 \mathbf{v}_{s}.
\end{aligned}
\end{equation}
Applying the chain rule through $t_{\text{mid}}$ and the coefficients, the gradients w.r.t. the $\textit{local}\,(l)$ $\textit{scaled}\,(s)$ coordinates are:
\begin{equation}
\label{eq:tmid->vs}
\begin{aligned}
\frac{\partial t_{\text{mid}}}{\partial \mathbf{v}_{s}} &=  \frac{\partial t_{\text{mid}}}{\partial b} \cdot \frac{\partial b}{\partial \mathbf{v}_{s}} + \frac{\partial t_{\text{mid}}}{\partial c} \cdot \frac{\partial c}{\partial \mathbf{v}_{s}}  \\&= -\frac{1}{a} \cdot \mathbf{d}_{s}, \\
\frac{\partial t_{\text{mid}}}{\partial \mathbf{d}_{s}} &=  \frac{\partial t_{\text{mid}}}{\partial a} \cdot \frac{\partial a}{\partial \mathbf{d}_{s}} + \frac{\partial t_{\text{mid}}}{\partial b} \cdot \frac{\partial b}{\partial \mathbf{d}_{s}} \\&= \frac{b}{a^2} \cdot \mathbf{d}_{s} -\frac{1}{a} \cdot \mathbf{v}_{s}.
\end{aligned}
\end{equation}

\noindent\textbf{Gradients w.r.t. Scale $\mathbf{s}$}.
The $\textit{local}\,(l)$ $\textit{scaled}\,(s)$ coordinates are defined as $\mathbf{v}_{s} = \mathbf{v}_{l} \oslash \mathbf{s}$ and $\mathbf{d}_{s} = \mathbf{d}_{l} \oslash \mathbf{s}$, where $\oslash$ denotes element-wise division. The gradients with respect to the scale $\mathbf{s}$ are computed as:
\begin{equation}
\label{eq:vs->s}
\frac{\partial (\mathbf{v}_s,\,\mathbf{d}_s)}{\partial \mathbf{s}} =   -\mathbf{v}_{l,k} \oslash \mathbf{s}  -\mathbf{d}_{l,k} \oslash \mathbf{s}.
\end{equation}

\noindent\textbf{Gradients w.r.t. Position $\boldsymbol{\mu}$ and Rotation $\mathbf{R}$}.
The local coordinates before scaling are obtained by rotating the world-space vectors into the Gaussian's local frame: $\mathbf{v}_{l} = \mathbf{R}^\top \mathbf{v}$ and $\mathbf{d}_{l} = \mathbf{R}^\top \mathbf{d}$, where $\mathbf{v} = \mathbf{o} - \boldsymbol{\mu}$.
The gradient with respect to the position $\boldsymbol{\mu}$ is
\begin{equation}
\label{eq:vs->mu}
\frac{\partial d_i}{\partial \boldsymbol{\mu}} = -\mathbf{R} \left( \frac{\partial d_i}{\partial \mathbf{v}_{l}} \oslash \mathbf{s} \right).
\end{equation}
The gradient with respect to the rotation matrix $\mathbf{R}$ is accumulated from both $\mathbf{v}$ and $\mathbf{d}$:
\begin{equation}
\label{eq:vs->r}
\frac{\partial d_i}{\partial \mathbf{R}} = \left( \frac{\partial d_i}{\partial \mathbf{v}_{l}} \oslash \mathbf{s} \right) \mathbf{v}^\top + \left( \frac{\partial d_i}{\partial \mathbf{d}_{l}} \oslash \mathbf{s} \right) \mathbf{d}^\top.
\end{equation}
Finally, the gradient $\frac{\partial d_i}{\partial \mathbf{R}}$ is converted to a gradient $\frac{\partial \mathbf{R}}{\partial \mathbf{q}}$ with respect to the quaternion $\mathbf{q}$ using the standard adjoint-sensitive conversion, ensuring the constraint of staying on the $S^3$ manifold through tangent space projection.

\subsection{Normal Estimation from Depth}\label{sec:normal_render}
Following the rendering of the depth map $\tilde{\mathbf{D}}$, we estimate a corresponding surface normal map $\tilde{\mathbf{N}}$ in a post-processing step. In this supplementary section, we will elaborate on two key operations of~\myalgref{alg:normal_estimation}: backprojecting depth pixels and orienting towards the view direction.

\noindent\textbf{Depth Backprojection to 3D Points}.
Each pixel $p=(u, v)$ with depth value $\tilde{\mathbf{D}}(u, v)$ is backprojected to a 3D point $\mathbf{P}_{\text{cam}}$ in the camera coordinate system using the intrinsic camera parameters:
\begin{equation}
\mathbf{P}_{\text{world}} = \begin{bmatrix} \mathbf{R}_{\text{cam}}^{\text{world}}
  & \mathbf{t}_{\text{cam}}^{\text{world}}\\\mathbf{0}^\top
  &1
\end{bmatrix} \begin{bmatrix}
\frac{(u - c_x)}{f_x} \cdot \tilde{\mathbf{D}}(u, v) \\
\frac{(v - c_y)}{f_y} \cdot \tilde{\mathbf{D}}(u, v) \\
\tilde{\mathbf{D}}(u, v) \\
1
\end{bmatrix},
\end{equation}
where $c_x = W/2$, $c_y = H/2$ are the coordinates of the principal point, and $f_x$, $f_y$ are the focal lengths.

\noindent\textbf{View Direction Correction}. Finally, the normal's orientation is corrected to face the camera. The normal is flipped if it points away from the camera, determined by the sign of the dot product between the normalized view direction $\mathbf{d}_{\text{view}}$ and the normal:
\begin{equation}
\label{eq:direction_correction}
\tilde{\mathbf{N}}(u, v) \leftarrow \begin{cases}
\tilde{\mathbf{N}}(u, v), & \text{if } \tilde{\mathbf{N}}(u, v) \cdot \mathbf{d}_{\text{view}} \leq 0 \\
-\tilde{\mathbf{N}}(u, v), & \text{otherwise}
\end{cases},
\end{equation}
where $\mathbf{d}_{\text{view}} = \frac{\mathbf{C} - \mathbf{P}_{\text{world}}}{\|\mathbf{C} - \mathbf{P}_{\text{world}}\|}$ is the normalized vector from the 3D point $\mathbf{P}_{\text{world}}$ to the camera center $\mathbf{C}$. This ensures all normals adhere to a consistent front-facing convention.

\begin{algorithm}[t]
\caption{Backward Propagation from Normals to Depth}
\label{alg:normal_gradient}
\textbf{Input:} $\frac{\partial \mathcal{L}_{normal}}{\partial \tilde{\mathbf{N}}}$,\,\texttt{STEP\_1},\,\texttt{STEP\_2},\,$\lambda$\\
\textbf{Output:} $\frac{\partial \mathcal{L}_{normal}}{\partial \tilde{\mathbf{D}}}$\\
\For{each pixel $p=(u, v)$ in parallel}{
    \textbf{1. View-Facing Orientation Flip}~\myeqref{equ:v_flip}$\gets\frac{\partial \mathcal{L}_{normal}}{\partial \tilde{\mathbf{N}}{(u, v)}}$ \\
    \textbf{2. Grad Normalization}~\myeqref{equ:grad_norm}\\
    \textbf{3. Grad Fusion}~\myeqref{equ:grad_fuse}$\gets \lambda$\\
    \textbf{4. Compute Multi-Scale Gradient} \\
    \Indp
    \For{$\texttt{STEP\_X} \in \{\texttt{STEP\_1}, \texttt{STEP\_2}\}$}{\textbf{Gradient Atom Addition}~\myeqref{equ:grad_finite1}~\myeqref{equ:grad_finite} \\
    }
    \Indm
    \textbf{5. Backprojection Gradient}~\myeqref{equ:grad_back} \\
    $\frac{\partial \mathcal{L}_{normal}}{\partial \tilde{\mathbf{D}}{(u, v)}}\gets$~\myeqref{equ:ld_all}
}
\end{algorithm}

\subsection{Backward Propagation from Normals to Depth}\label{sec:normal_backward}
To enable optimization via losses defined on the rendered normal map $\tilde{\mathbf{N}}$, the gradients must be backpropagated through the normal estimation pipeline to the depth map $\tilde{\mathbf{D}}$ and subsequently to the underlying 3D Gaussian parameters. 
Overall gradient backpropagation pipeline is shown in~\myalgref{alg:normal_gradient}.
The gradient of the loss $\mathcal{L}_{normal}$ with respect to the depth map $\tilde{\mathbf{D}}$ is propagated by
\begin{equation}
\label{equ:ld_all}
\begin{aligned}
\frac{\partial \mathcal{L}_{normal}}{\partial \tilde{\mathbf{D}}(u,v)} = \frac{\partial \mathcal{L}_{normal}}{\partial \tilde{\mathbf{N}}(u,v)} \cdot \frac{\partial \tilde{\mathbf{N}}(u,v)}{\partial \mathbf{n}_{\text{fused}}^{\text{(orient)}}} \cdot\frac{\partial \mathbf{n}_{\text{fused}}^{\text{(orient)}}}{\partial \mathbf{n}_{\text{fused}}} \\ \cdot \frac{\partial \mathbf{n}_{\text{fused}}}{\partial (\mathbf{n}_1, \mathbf{n}_2)}  \cdot \left( \frac{\partial \mathbf{n}_1}{\partial {\mathbf{P}_{\text{world}}^{n_1}}} + \frac{\partial \mathbf{n}_2}{\partial {\mathbf{P}_{\text{world}}^{n_2}}} \right) \cdot \frac{\partial {\mathbf{P}_{\text{world}}}}{\partial \tilde{\mathbf{D}}(u,v)},
\end{aligned}
\end{equation}
where $\mathbf{P}_{\text{world}}^{n_1}$ and $\mathbf{P}_{\text{world}}^{n_2}$ represent the sets of 3D points in the neighborhoods used to compute $\mathbf{n}_1$ and $\mathbf{n}_2$, respectively. 

\noindent\textbf{Cosine Similarity w.r.t. Normal.}
For encouraging geometric consistency~\myeqref{eq:csloss} is the negative cosine similarity between the predicted normal $\tilde{\mathbf{N}}$ and a target normal $\tilde{\mathbf{N}}_{\text{target}}$:
\begin{equation}
\mathcal{L}_{\text{normal}} =1 -\frac{1}{|\Omega|} \sum_{(u,v) \in \Omega} \tilde{\mathbf{N}}(u, v) \cdot \tilde{\mathbf{N}}_{\text{target}}(u, v),
\end{equation}
where $\Omega$ denotes the set of pixels. The gradient of this loss with respect to the predicted normal at a specific pixel is straightforward:
\begin{equation}\label{eq:gradientn}
\frac{\partial \mathcal{L}_{normal}}{\partial \tilde{\mathbf{N}}(u,v)} = -\frac{1}{|\Omega|} \tilde{\mathbf{N}}_{\text{target}}(u, v).
\end{equation}
This gradient $\frac{\partial \mathcal{L}_{normal}}{\partial \tilde{\mathbf{N}}}$ is the seed that initiates the backward pass through the normal estimation pipeline.

\noindent\textbf{View-facing Orientation Flip.} If the normal is flipped in the forward pass to face the viewer (defined by~\myeqref{eq:direction_correction}), the incoming gradient is correspondingly flipped:
\begin{equation}
\label{equ:v_flip}
\frac{\partial \mathcal{L}_{normal}}{\partial \mathbf{n}_{\text{fused}}^{\text{(orient)}}} =
\begin{cases}
-\frac{\partial \mathcal{L}_{normal}}{\partial \tilde{\mathbf{N}}(u,v)}, & \text{if flipped} \\
\frac{\partial \mathcal{L}_{normal}}{\partial \tilde{\mathbf{N}}(u,v)}, & \text{otherwise}
\end{cases},
\end{equation}

\noindent\textbf{Normalization.} The gradient is then backpropagated through the normalization operation $\tilde{\mathbf{N}} = \mathbf{n}_{\text{fused}} / \|\mathbf{n}_{\text{fused}}\|$. The gradient w.r.t. the unnormalized vector $\mathbf{n}_{\text{fused}}$ is given by:
\begin{equation}
\label{equ:grad_norm}
\frac{\partial \mathbf{n}_{\text{fused}}^{\text{(orient)}}}{\partial \mathbf{n}_{\text{fused}}} = \frac{\left( \mathbf{I} - \tilde{\mathbf{N}}(u,v) \tilde{\mathbf{N}}(u,v)^\top \right)}{\|\mathbf{n}_{\text{fused}}\|} .
\end{equation}

\noindent\textbf{Normal Fusion.} The gradient is distributed to the two constituent normal estimates $\mathbf{n}_1$ and $\mathbf{n}_2$ from which $\mathbf{n}_{\text{fused}} = \lambda \mathbf{n}_1 + (1-\lambda)\mathbf{n}_2$ is computed. If their directions were inconsistent in the forward pass ($\mathbf{n}_1 \cdot \mathbf{n}_2 < 0$), the sign of the gradient for $\mathbf{n}_2$ is flipped:
\begin{equation}
\label{equ:grad_fuse}
\begin{aligned}
\frac{\partial \mathbf{n}_{\text{fused}}}{\partial \mathbf{n}_1}= \lambda ,\,
\frac{\partial \mathbf{n}_{\text{fused}}}{\partial \mathbf{n}_2}= (1-\lambda) \cdot \text{sign}(\mathbf{n}_1 \cdot \mathbf{n}_2).
\end{aligned}
\end{equation}

\noindent\textbf{Finite Differences to 3D Points.} The core of the backward pass involves computing how the estimated normals $\mathbf{n_1}$ and $\mathbf{n_2}$ change with respect to the 3D points $\mathbf{P}_{\text{world}}$ in their finite-difference neighborhood. For simplicity, we will take $\mathbf{n_X}\in \{\mathbf{n_1}, \mathbf{n_2}\}$, $\texttt{STEP\_X}\in \{\texttt{STEP\_1}, \texttt{STEP\_2}\}$ as an example for the subsequent derivation. For a normal estimate $\mathbf{n_X}$ calculated from points $\mathbf{P}$ in a $\texttt{STEP\_X} \times \texttt{STEP\_X}$ window, the gradient w.r.t. each involved point $\mathbf{P}_{\text{world}}^{\mathbf{n_X}}$ is derived from the cross product operation.

From~\myalgref{alg:normal_estimation}, $\mathbf{v}_x^{\mathbf{n_X}} = \mathbf{P}_{\text{world}}^{\mathbf{n_X}}(u+\texttt{STEP\_X},v) - \mathbf{P}_{\text{world}}^{\mathbf{n_X}}(u-\texttt{STEP\_X},v)$ and $\mathbf{v}_y^{\mathbf{n_X}} = \mathbf{P}_{\text{world}}^{\mathbf{n_X}}(u,v+\texttt{STEP\_X}) - \mathbf{P}_{\text{world}}^{\mathbf{n_X}}(u,v-\texttt{STEP\_X})$), the normal is $\mathbf{n_X} = \mathbf{v}_x^{\mathbf{n_X}} \times \mathbf{v}_y^{\mathbf{n_X}}$. The gradients of $\mathbf{n_X}$ w.r.t. the vectors $\mathbf{v}_x^{\mathbf{n_X}}$ and $\mathbf{v}_y^{\mathbf{n_X}}$ are:
\begin{equation}
\label{equ:grad_finite1}
\frac{\partial \mathbf{n_X}}{\partial \mathbf{v}_x^{\mathbf{n_X}}} = \mathbf{v}_y^{\mathbf{n_X}},\,
\frac{\partial \mathbf{n_X}}{\partial \mathbf{v}_y^{\mathbf{n_X}}} = \mathbf{v}_x^{\mathbf{n_X}}.
\end{equation}
These gradients are then distributed to the specific points that contributed to $\mathbf{v}_x^{\mathbf{n_X}}$ and $\mathbf{v}_y^{\mathbf{n_X}}$:
\begin{equation}
\label{equ:grad_finite}
\begin{aligned}
\frac{\partial \mathbf{n_X}}{\partial \mathbf{P}_{\text{world}}^{\mathbf{n_X}}(u+\texttt{STEP\_X},v)} \mathrel{+}=  \mathbf{v}_x^{\mathbf{n_X}},\\
\frac{\partial \mathbf{n_X}}{\partial \mathbf{P}_{\text{world}}^{\mathbf{n_X}}(u-\texttt{STEP\_X},v)} \mathrel{-}= \mathbf{v}_x^{\mathbf{n_X}},\\
\frac{\partial \mathbf{n_X}}{\partial \mathbf{P}_{\text{world}}^{\mathbf{n_X}}(u,v+\texttt{STEP\_X})} \mathrel{+}=  \mathbf{v}_y^{\mathbf{n_X}},\\
\frac{\partial \mathbf{n_X}}{\partial \mathbf{P}_{\text{world}}^{\mathbf{n_X}}(u,v-\texttt{STEP\_X})} \mathrel{-}= \mathbf{v}_y^{\mathbf{n_X}}.
\end{aligned}
 \end{equation}
This process is repeated for \textit{both} normal estimates $\mathbf{n}_1$ and $\mathbf{n}_2$ and their respective neighborhoods $\mathbf{P}_{\text{world}}^{n_1}$ and $\mathbf{P}_{\text{world}}^{n_2}$, and the gradients are accumulated atomically.

\noindent\textbf{3D Points w.r.t Depth.}
The Jacobian matrix $\frac{\partial \mathbf{P}_{\text{world}}}{\partial \tilde{\mathbf{D}}(u,v)}$ is derived from the backprojection and transformation chain $\mathbf{P}_{\text{world}} = \mathbf{R}^{\text{world}}_{\text{cam}} \mathbf{P}_{\text{cam}}(\tilde{\mathbf{D}}(u,v))$:
\begin{equation}
\label{equ:grad_back}
\frac{\partial \mathbf{P}_{\text{world}}}{\partial \tilde{\mathbf{D}}(u,v)} = \mathbf{R}^{\text{world}}_{\text{cam}}\frac{\partial \mathbf{P}_{\text{cam}}}{\partial \tilde{\mathbf{D}}(u,v)} = \mathbf{R}^{\text{world}}_{\text{cam}}
\begin{bmatrix}
(u - c_x)/f_x \\
(v - c_y)/f_y \\
1 \\
0
\end{bmatrix}.
\end{equation}

\mysubsection{Differentiable Pruning}\label{sec:diff_prune}

\noindent\textbf{Gradient Propagation.} During backpropagation of $k$, the gradient can be propagated to the gradient factor $k$ through the rendering equation. The chain rule for the gradient of the loss with respect to a specific $k_i$ can be conceptually expressed as
\begin{equation}
\label{eq:gradientk}
\frac{\partial \mathcal{L}_K(u, v)}{\partial k_i} = \frac{\partial \mathcal{L}_K(u, v)}{\partial \tilde{\mathbf{K}}(u, v)} \cdot \frac{\partial \tilde{\mathbf{K}}(u, v)}{\partial k_i},
\end{equation}
where $\frac{\partial \tilde{\mathbf{K}}(u, v)}{\partial k_i}$ can be derived from~\myeqref{eq:t_render} in main text.

\noindent\textbf{Detailed Analysis of optimal state $k=1$.}
The gradient factor $k$ serves as a core indicator for Gaussian pruning. Its deviation from 1 directly reflects the optimization state of Gaussians, and the underlying causes along with pruning rationale are elaborated as follows.

\noindent$\bullet$ \textbf{Gaussians with $k > 1$: Anomalies from Gradient Fluctuations Caused by Data Noise.}  
Gaussians with extremely large $k$-values are induced by gradient anomalies during multi-view optimization. During the rendering of specific views, data noise such as sensor noise and texture inconsistencies at scene edges leads to significant fluctuations in pixel gradients. These anomalous gradients propagate to the opacity $\alpha$ through backpropagation, causing $\alpha$ to deviate from its optimal value. This deviation results in unexpected optimization under specific view. Such abnormal $k$-values are usually accompanied by extreme Gaussian parameters including position, scale as shown in~\myfigref{fig:prune_vis} in the main text. 

\noindent$\bullet$ \textbf{Gaussians with $k < 1$: Redundancy Due to Insufficient Rendering Participation.}  
Gaussians with small $k$-values can be identified based on long-term optimization, especially considering the periodic $k$-reset strategy. During training, we periodically reset the $k$-value of all Gaussians to 0.9. This initial value is designed to provide a starting point for optimization driven by $\mathcal{L}_K$. The expectation is that after multiple optimization iterations, well-optimizing Gaussians will adjust their $k$-values from 0.9 toward 1. The ideal state is when $k = 1$, which makes the rendering formula of $\tilde{\mathbf{K}}$ satisfy 
\begin{equation}
\tilde{\mathbf{K}}(u, v) = \sum_{i\in \mathcal{N}} \alpha_i \cdot 1 \cdot \prod_{j=1}^{i-1}(1 - \alpha_j) = 1.
\end{equation}
If a Gaussian’s $k$-value remains near 0.9 or changes minimally after a long optimization horizon, it indicates the Gaussian has not deeply participated in the rendering of any view. Specifically, it lacks sufficient effective alpha-rendering contributions to pixels across views. This situation will occur because the Gaussian is located in unobservable regions such as persistent occlusions. Such Gaussians are redundant, and their exclusion does not affect the accuracy of scene reconstruction as they contribute little to the rendering of $\tilde{\mathbf{K}}$ or RGB images.

\noindent\textbf{Pruning Mechanism Based on $k$.}  
In summary, well-optimized Gaussians can adjust their $k$-values toward 1 under the constraint of $\mathcal{L}_K$, and their $\alpha$ and geometric parameters can well match the scene without the need for extreme $k$-compensation. In contrast, Gaussians with $|k - 1| > \mathcal{T}_k$ are either anomalous with $k > 1$ or redundant with $k < 1$. By pruning these Gaussians, we can effectively reduce the total number of Gaussians while preserving the quality of scene reconstruction, thereby improving the overall computational efficiency.

\begin{algorithm}[!htb]
\caption{Tile-based Backward Gradient Propagation Pipeline}
\label{alg:backward_render}
\KwIn{Forward rendering results: $\tilde{\mathbf{D}}$, $\tilde{\mathbf{N}}$, $\tilde{\mathbf{O}}$, $\tilde{\mathbf{K}}$\; Loss gradients: $\frac{\partial \mathcal{L}_{depth}}{\partial \tilde{\mathbf{D}}}$, $\frac{\partial \mathcal{L}_{normal}}{\partial \tilde{\mathbf{N}}}$, $\frac{\partial \mathcal{L}_{seg}}{\partial \tilde{\mathbf{O}}}$, $\frac{\partial \mathcal{L}_{K}}{\partial \tilde{\mathbf{K}}}$}
\KwOut{Parameter gradients: $\frac{\partial \mathcal{L}}{\partial \boldsymbol{\mu}}$, $\frac{\partial \mathcal{L}}{\partial \mathbf{R}}$, $\frac{\partial \mathcal{L}}{\partial \mathbf{s}}$, $\frac{\partial \mathcal{L}}{\partial \mathbf{o}}$, $\frac{\partial \mathcal{L}}{\partial K}$}

\ForPar{each pixel $(x, y)$}{
    	\tcp{Restore Saved Forward Values}
    Restore: $T \gets T_{\text{final}}$\;
    \For{each Gaussian $i$ intersecting pixel $pix$ (in reverse depth order)}{
    	$...$
    	
        $T \gets T / (1 - \alpha_i)$ \;
        $\mathbf{w_\alpha} \gets \alpha_i \cdot T$\;
        
        \tcp{Compute Gradients for Rendered Attributes}
        \For{each $ch$ in $C_o$}{
        $\frac{\partial \mathcal{L}_{all}}{\partial \mathbf{o}_{i[ch]}} \gets \underbrace{\frac{\partial \mathcal{L}_{seg}}{\partial \tilde{\mathbf{O}}_{[ch]}}}_{~\myeqref{eq:gradiento}} \cdot \mathbf{w_\alpha}$\;
       	}
        $\frac{\partial \mathcal{L}_{all}}{\partial k_i} \gets \underbrace{\frac{\partial \mathcal{L}_{K}}{\partial \tilde{\mathbf{K}}}}_{~\myeqref{eq:gradientk}} \cdot \mathbf{w_\alpha}$\;
        
        \tcp{Depth Gradient Propagation}
        $\frac{\partial \mathcal{L}_{all}}{\partial d_i} \gets(\underbrace{\frac{\partial \mathcal{L}_{normal}}{\partial \tilde{\mathbf{N}}}}_{~\myeqref{eq:gradientn}} + \frac{\partial \mathcal{L}_{depth}}{\partial \tilde{\mathbf{D}}}) \cdot \mathbf{w_\alpha}$\;
        
        \tcp{Propagate Depth Gradient to Ellipsoid Parameters}
        $\frac{\partial \mathcal{L}_{all}}{\partial \mathbf{s}} = \frac{\partial \mathcal{L}_{all}}{\partial d_i}\cdot\underbrace{\frac{\partial d_i}{\partial \mathbf{s}}}_{\myeqref{eq:vs->s}}$\;
        $\frac{\partial \mathcal{L}_{all}}{\partial \boldsymbol{\mu}} = \frac{\partial \mathcal{L}_{all}}{\partial d_i}\cdot\underbrace{\frac{\partial d_i}{\partial \boldsymbol{\mu}}}_{\myeqref{eq:vs->mu}}$\;
        $\frac{\partial \mathcal{L}_{all}}{\partial \mathbf{q}} = \frac{\partial \mathcal{L}_{all}}{\partial d_i}\cdot\underbrace{\frac{\partial d_i}{\partial \mathbf{R}}}_{\myeqref{eq:vs->r}}\cdot\frac{\partial \mathbf{R}}{\partial \mathbf{q}}$\;

		$...$
    }
}
\end{algorithm}

\subsection{Tile-Based Gradient Backpropagation}\label{sec:backward_render}
In the previous sections, we obtain the losses for the semantic logits, depth, normal, and gradient factor map with respect to their respective ground truths, which are given by \(\frac{\partial \mathcal{L}}{\partial \tilde{\mathbf{O}}}\), \(\frac{\partial \mathcal{L}}{\partial \tilde{\mathbf{D}}}\), \(\frac{\partial \mathcal{L}}{\partial \tilde{\mathbf{N}}}\), and \(\frac{\partial \mathcal{L}}{\partial \tilde{\mathbf{K}}}\). Corresponding to~\myalgref{alg:forward_render}, we optimize the Gaussian parameters through the backward propagation of gradients via the alpha rendering process. The backward propagation pipeline is illustrated in~\myalgref{alg:backward_render}. We omit the gradient propagation process for RGB (indicated as $...$), which is consistent with the original implementation~\cite{3dgs}.

\begin{table*}[!htb]
\centering
\caption{
Quantitative comparison of RGB rendering on widely used real and simulated datasets. Some metrics ($^*$) are extracted from the original data in respective papers. The results ranked from best to worst are highlighted as \colorbox[HTML]{c0e2ca}{first}, \colorbox[HTML]{fff5b3}{second}, and \colorbox[HTML]{ffd9b3}{third}.
}
\resizebox{\linewidth}{!}{
\begin{tabular}{clcccccccccccc}
\hline
\multicolumn{2}{c}{Dataset}          & \multicolumn{3}{c}{Mip-NeRF360}                                                               & \multicolumn{3}{c}{Tanks \& Temples}                                                          & \multicolumn{3}{c}{Deep Blending}                                                             & \multicolumn{3}{c}{Replica}                                                                   \\
\multicolumn{2}{c}{Method $|$ Metrics} & PSNR   ↑                       & SSIM ↑                         & LPIPS       ↓                  & PSNR     ↑                     & SSIM      ↑                    & LPIPS        ↓                 & PSNR     ↑                     & SSIM        ↑                  & LPIPS       ↓                  & PSNR          ↑                & SSIM     ↑                     & LPIPS              ↓           \\ \hline
\multicolumn{2}{c}{iNPG*~\cite{muller2022instant}}            & 26.43                         & 0.725                         & 0.339                         & 21.72                         & 0.723                         & 0.330                         & 23.62                         & 0.797                         & 0.423                         & -                             & -                             & -                             \\
\multicolumn{2}{c}{Mip-NeRF*~\cite{barron2022mip}}        & \cellcolor[HTML]{FFD9B3}29.23 & 0.844                         & 0.207                         & 22.22                         & 0.759                         & 0.257                         & 29.40                         & 0.901                         & 0.245                         & -                             & -                             & -                             \\
\multicolumn{2}{c}{RaySplats*~\cite{raysplat}}       & 27.31                         & 0.846                         & 0.237                         & 22.20                         & 0.829                         & 0.202                         & 29.57                         & 0.900                         & 0.320                         & -                             & -                             & -                             \\
\multicolumn{2}{c}{GeoGaussian~\cite{li2024geogaussian}}      & 26.26                         & 0.813                         & 0.219                         & 20.00                         & 0.802                         & 0.261                         & 26.07                         & 0.873                         & 0.310                         & 24.55                         & 0.946                         & 0.114                         \\
\multicolumn{2}{c}{3DGS~\cite{3dgs}}             & 28.69                         & \cellcolor[HTML]{FFF5B3}0.870 & \cellcolor[HTML]{FFF5B3}0.182 & 23.14                         & 0.841                         & 0.183                         & 29.41                         & \cellcolor[HTML]{FFD9B3}0.903 & \cellcolor[HTML]{FFF5B3}0.243 & 36.50                         & 0.940                         & 0.120                         \\
\multicolumn{2}{c}{Scaffold-GS~\cite{scaffoldgs}}      & 28.84                         & 0.848                         & 0.220                         & \cellcolor[HTML]{FFF5B3}23.96 & \cellcolor[HTML]{FFD9B3}0.853 & \cellcolor[HTML]{FFD9B3}0.177 & \cellcolor[HTML]{C0E2CA}30.21 & \cellcolor[HTML]{FFF5B3}0.906 & 0.254                         & \cellcolor[HTML]{FFF5B3}37.45 & \cellcolor[HTML]{FFD9B3}0.951 & \cellcolor[HTML]{FFD9B3}0.090 \\
\multicolumn{2}{c}{2DGS~\cite{huang20242d}}             & 27.34                         & 0.814                         & 0.234                         & 22.89                         & 0.833                         & 0.206                         & 27.96                         & 0.873                         & 0.281                         & 36.47                         & 0.943                         & 0.107                         \\
\multicolumn{2}{c}{Normal-GS~\cite{normalgs}}        & \cellcolor[HTML]{FFF5B3}29.34 & \cellcolor[HTML]{FFD9B3}0.869 & \cellcolor[HTML]{FFD9B3}0.194 & \cellcolor[HTML]{C0E2CA}24.22 & \cellcolor[HTML]{FFF5B3}0.854 & \cellcolor[HTML]{FFF5B3}0.174 & \cellcolor[HTML]{FFF5B3}30.19 & \cellcolor[HTML]{C0E2CA}0.910 & \cellcolor[HTML]{FFD9B3}0.252 & \cellcolor[HTML]{C0E2CA}38.19 & \cellcolor[HTML]{C0E2CA}0.977 & \cellcolor[HTML]{C0E2CA}0.043 \\ \hline
\multicolumn{2}{c}{Ours}             & \cellcolor[HTML]{C0E2CA}30.10 & \cellcolor[HTML]{C0E2CA}0.887 & \cellcolor[HTML]{C0E2CA}0.148 & \cellcolor[HTML]{FFD9B3}23.57 & \cellcolor[HTML]{C0E2CA}0.860 & \cellcolor[HTML]{C0E2CA}0.168 & \cellcolor[HTML]{FFD9B3}29.66 & 0.899                         & \cellcolor[HTML]{C0E2CA}0.231 & \cellcolor[HTML]{FFD9B3}36.68 & \cellcolor[HTML]{FFF5B3}0.962 & \cellcolor[HTML]{FFF5B3}0.083 \\ \hline
\end{tabular}}
\label{tab:rgb_rendering}
\end{table*}

\begin{figure*}[tbp]
\centering
\includegraphics[width=\linewidth]{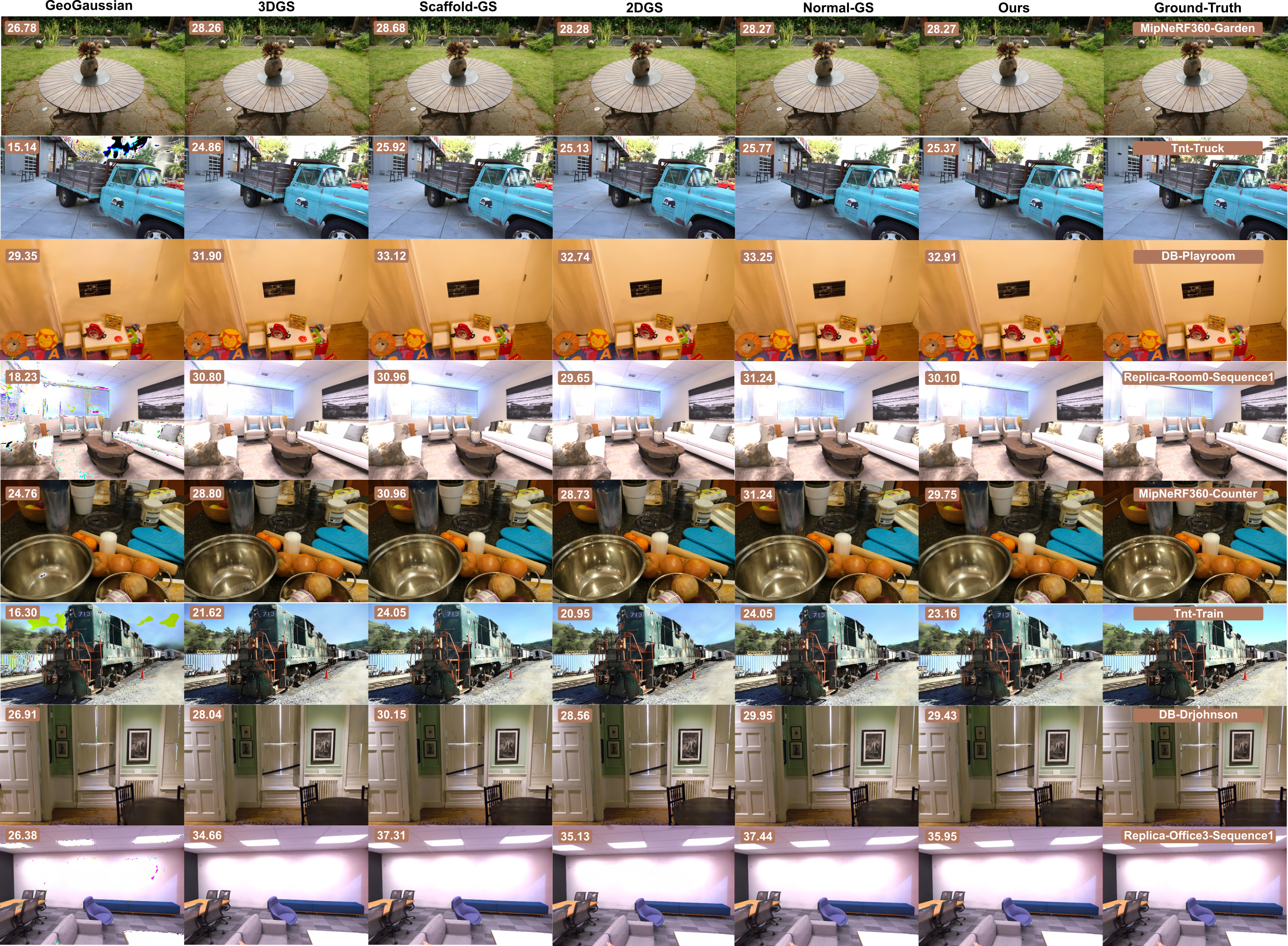}
\caption{Qualitative comparison results on NVS demonstrate that our RGB rendering achieves better performance in certain scenarios. The top-left corner of the image displays the PSNR value computed against the ground truth image.}
\label{fig:rgb_rendering}
\vspace{-1em}
\end{figure*}


\mysection{More Experiments}
\mysubsection{RGB Rendering Evaluation}
\label{sec:rgb_rendering}
To validate the effectiveness of our method in providing valuable information for NVS, we compare its performance on novel view RGB rendering with state-of-the-art frameworks~\cite{barron2022mip}~\cite{muller2022instant}~\cite{yu2021plenoxels}~\cite{3dgs}~\cite{scaffoldgs}~\cite{li2024geogaussian}. All methods utilize their provided original code and unified original data. The qualitative results of the RGB rendering are shown in~\myfigref{fig:rgb_rendering}, and the quantitative comparisons are presented in~\mytabref{tab:rgb_rendering}. It can be observed that our method achieves state-of-the-art results in novel view synthesis for certain scenes. By incorporating geometric constraints into the framework, our algorithm achieves SOTA geometric performance while maintaining RGB reconstruction quality and significantly improving memory and rendering efficiency.

\begin{table*}[t]
\centering
\caption{Quantitative comparisons of the ablation experiments on each module.
}
\resizebox{0.99\linewidth}{!}{
\begin{tabular}{clcccccccccccc}
\hline
\multicolumn{2}{c}{Datasets}               & \multicolumn{6}{c}{Room0-Sequence1}                & \multicolumn{6}{c}{Office3-Sequence3}              \\
\multicolumn{2}{c}{Settings $|$ Metrics}       & PSNR↑   & Abs.Rel.↓ & CosSimi↑     & mIoU↑   & Count↓  & FPS↑ & PSNR↑   & Abs.Rel.↓ & CosSimi↑     & mIoU↑   & Count↓  & FPS↑   \\ \hline
\multicolumn{2}{c}{RayCasting-Point (All)} & 31.341 & 0.0062   & 0.855 & 0.607 & 170859 & 183.8 & 36.892 & 0.0208   & 0.931 & 0.534 & 178905 & 184.8 \\
\multicolumn{2}{c}{Center Point}           & 31.800 & 0.0071   & 0.849 & 0.607 & 174177 & 181.7 & 36.873 & 0.0204   & 0.932 & 0.534 & 179128 & 187.8 \\ \hline
\multicolumn{2}{c}{w/o prune}              & 31.509 & 0.0070   & 0.851 & 0.606 & 229487 & 156.3 & 37.100 & 0.0315   & 0.928 & 0.534 & 237914 & 165.5 \\
\multicolumn{2}{c}{w/prune $\mathcal{T}_k$=0.2}         & 31.273 & 0.0068   & 0.847 & 0.605 & 144185 & 204.1 & 36.925 & 0.0161   & 0.932 & 0.534 & 148167 & 189.7 \\ \hline
\multicolumn{2}{c}{w/o seg}                & 31.605 & 0.0226   & 0.837 & 0.021 & 174230 & 188.0 & 36.833 & 0.0225   & 0.929 & 0.024 & 184952 & 168.9 \\
\multicolumn{2}{c}{w/o normal}             & 32.750 & 0.0061   & 0.774 & 0.608 & 167494 & 206.5 & 37.488 & 0.00885  & 0.895 & 0.535 & 182019 & 203.8 \\
\multicolumn{2}{c}{w/o depth}              & 32.161 & 0.0184   & 0.850 & 0.607 & 171462 & 189.7 & 37.318 & 0.0985   & 0.925 & 0.539 & 176346 & 184.1 \\
\multicolumn{2}{c}{w/o depth\&normal}      & 33.691 & 0.0214   & 0.658 & 0.609 & 167198 & 202.6 & 37.449 & 0.1106   & 0.782 & 0.539 & 180747 & 189.7 \\ \hline
\end{tabular}}
\label{tab:sup:abla_all}
\end{table*}

\mysubsection{Statistical Outlier Analysis}
To quantify the quality of predicted point clouds relative to ground truth data in Replica~\cite{straub2019replica}, we employ a statistical approach based on Z-score analysis of point-to-point distances. This method allows us to identify and measure outliers in predicted point clouds with respect to the ground truth mesh points.

\noindent\textbf{Distance.}
For each point $p_i$ in the predicted point cloud $P$, we compute the minimum Euclidean distance to the ground truth point cloud $G$:
\begin{equation}
d_i = \min_{g \in G} |p_i - g|,
\end{equation}
where $|\cdot|$ denotes the Euclidean norm. This computation is efficiently implemented using a KD-tree data structure for nearest neighbor search.

\noindent\textbf{Statistical Measures.}
From the computed distances ${d_1, d_2, \ldots, d_n}$ where $n$ is the number of points in the predicted point cloud, we calculate the following statistical measures:
\begin{equation}
\mu_d = \frac{1}{n} \sum_{i=1}^{n} d_i,
\end{equation}
\begin{equation}
\sigma_d = \sqrt{\frac{1}{n} \sum_{i=1}^{n} (d_i - \mu_d)^2},
\end{equation}
where $\mu_d$ represents the \textbf{Mean} distance and $\sigma_d$ represents the standard deviation (\textbf{Std}) of distances.

For each distance $d_i$, we compute its Z-score, which measures how many standard deviations the distance is from the mean:
\begin{equation}
z_i = \frac{|d_i - \mu_d|}{\sigma_d}.
\end{equation}
Points are classified as outliers based on a threshold parameter $\tau$ ($\tau = 0.1$ in our experiments):
\begin{equation}
\mathcal{F}_i =\left\{\begin{matrix}
  1 & \text{if } z_i > \tau \\
0 & \text{otherwise}
\end{matrix}\right..
\end{equation}
The outlier ratio (\textbf{Radio}) is then calculated as the proportion of points classified as outliers:
\begin{equation}
\text{Radio} = \frac{1}{n} \sum_{i=1}^{n}\mathcal{F}_i.
\end{equation}
\noindent\textbf{Hausdorff Distance.}
In addition to the statistical outlier analysis, we compute the Hausdorff distance to measure the maximum deviation between point clouds:
\begin{equation}
H(G, P) = \max\left\{ \sup_{g \in G} \inf_{p \in P} |g - p|, \sup_{p \in P} \inf_{g \in G} |p - g| \right\}.
\end{equation}

\begin{figure*}[tbp]
\centering
\includegraphics[width=\linewidth]{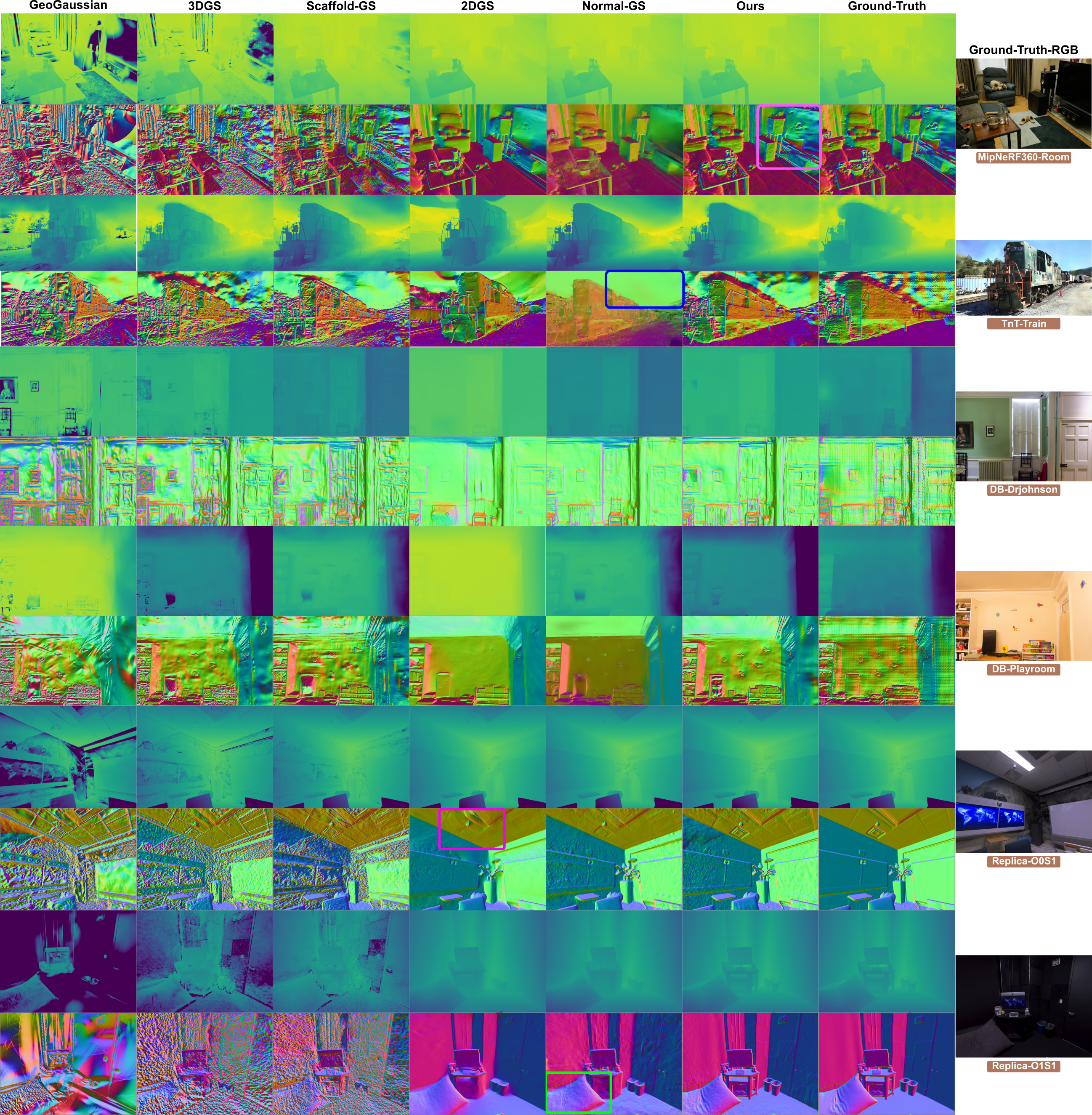}
\caption{Qualitative comparison of depth and normal estimation accuracy across different methods on several datasets.}
\label{fig:supp:depth_normal_rendering}
\vspace{-1em}
\end{figure*}

\mysubsection{More Results on Geometric Rendering}
We present additional comparative results of depth and surface normal predictions across multiple datasets in~\myfigref{fig:supp:depth_normal_rendering}, which demonstrate the superiority of our method in geometric rendering. Particularly in texture-sparse regions, where other methods fail to accurately reconstruct object surfaces, our approach consistently ensures high consistency among RGB, depth, and normal estimations.

It is worth noting that in the Mip-NeRF~\cite{barron2022mip}, Tanks \& Temples~\cite{knapitsch2017tanks}, and Deep Blending~\cite{hedman2018deep} datasets, the depth and normal information used for supervision is provided by Normal-GS~\cite{normalgs} and may still be inaccurate in certain scenarios. Nevertheless, our method achieves competitive results in both depth and normal estimation.

\mysubsection{Ablation on Algorithm Components}
We test the quantitative rendering quality and runtime analysis with and without the differentiable pruning module, as shown in~\mytabref{tab:sup:abla_all}. It is evident that our differentiable pruning module effectively reduces the number of Gaussians used without significantly compromising image rendering quality. thereby improving rendering FPS and reducing stray points in the infinite distance.

Furthermore, \mytabref{tab:sup:abla_all} demonstrates the performance gains achieved by our ray-casting based method on depth and normal estimation, as well as the impact of integrating various loss functions on their respective evaluation metrics. It can be observed that our ray-casting based approach enhances the algorithm's performance on geometric metrics. 

Unlike methods that optimize for a single modality, our focus lies in achieving a consistent representation of all modalities in the scene. We therefore strive for a balanced performance across all metrics, which ultimately leads to a more faithful reconstruction of the real world.

\mysection{Multimodal Viewer}
We also extend a visualization rendering software based on the \textit{SIBR\_viewer} and a CUDA rendering pipeline we developed. This tool supports real-time visualization of RGB, depth, normal, and semantic maps, achieving a rendering speed of up to 200 FPS. The corresponding viewer will also been open-sourced on GitHub.

\mysection{Limitations}
Real-world environments exhibit complex optical interactions including extensive light reflections and inter reflections. Image captured by a camera is the product of multi-path light propagation and global illumination effects. Our current approach does not explicitly model these radiometric phenomena such as light reflection and composition. Furthermore, this work focuses exclusively on supervised offline reconstruction of static scenes. Extending the framework to incorporate temporal coherence for dynamic scene modeling remains a direction for future work.